\documentclass[final,5p,times]{elsarticle}

\usepackage{graphicx}
\usepackage{epsfig}

\usepackage[ruled,vlined]{algorithm2e}

\usepackage{amssymb}
\usepackage{amsmath}

\usepackage{hyperref}
\usepackage{etoolbox}
\usepackage{multirow}
\usepackage{subfigure}
\usepackage{longtable}
\usepackage{tabu}
\usepackage{booktabs}
\usepackage{rotating}
\usepackage{balance}
\usepackage[usenames, dvipsnames]{color}
\newcommand{\hide}[1]{}

\newcommand{\dquote}[1]{``#1''}
\newcommand{\etal}{\textit{et al.}~}

\newtoggle{single}
\togglefalse{single}

\journal{Pattern Recognition}
\begin{document}

\begin{frontmatter}


\author[micc]{Tiberio Uricchio}
\ead{tiberio.uricchio@unifi.it}
\author[unipd]{Lamberto Ballan\corref{cor}}
\ead{lamberto.ballan@unipd.it}
\author[micc]{Lorenzo Seidenari}
\ead{lorenzo.seidenari@unifi.it}
\author[micc]{Alberto Del Bimbo}
\ead{alberto.delbimbo@unifi.it}

\cortext[cor]{Corresponding author. A major part of this work has been done while the author was on an EU Marie Curie Fellowship at Stanford University and Univ. of Florence.}
\address[micc]{Media Integration and Communication Center (MICC), Universit\`a degli Studi di Firenze, Viale Morgagni 65, 50134 Firenze, Italy}
\address[unipd]{Department of Mathematics \dquote{Tullio Levi-Civita}, Universit\`a degli Studi di Padova, Via Trieste 63, 35121 Padova, Italy\vspace{-7pt}}


\title{Automatic Image Annotation via Label Transfer in the Semantic Space}

\begin{abstract}
Automatic image annotation is among the fundamental problems in computer vision and pattern recognition, and it is becoming increasingly important in order to develop algorithms that are able to search and browse large-scale image collections.  
In this paper, we propose a label propagation framework based on Kernel Canonical Correlation Analysis (KCCA), which builds a latent \emph{semantic space} where correlation of visual and textual features are well preserved into a semantic embedding.
The proposed approach is robust and can work either when the training set is well annotated by experts, as well as when it is noisy such as in the case of user-generated tags in social media.
We report extensive results on four popular datasets. Our results show that our KCCA-based framework can be applied to several state-of-the-art label transfer methods to obtain significant improvements.
Our approach works even with the noisy tags of social users, provided that appropriate denoising is performed.
Experiments on a large scale setting show that our method can provide some benefits even when the semantic space is estimated on a subset of training images.
\end{abstract}


\end{frontmatter}



\section{Introduction}\label{sec:intro}
A lot of modern applications require image annotation to search, access and navigate the huge amount of visual data stored in personal collections or shared online.
Whenever you want to retrieve photos from a particular concert, recall that pleasant summer day in which you napped on your comfortable hammock or look up a person, it is automatic image annotation that enables a plethora of useful applications. The exponential growth of media on sharing platforms, such as Flickr or Facebook, has led to the availability of a huge quantity of images that are enjoyed by millions of people. In such a huge sea of data, it is indispensable to teach computers to correctly label the visual content and help us search and browse image collections.

In this paper, we tackle the challenging task of automatic image annotation.
Given an image, we want to assign a set of relevant labels by taking into account image appearance and eventually some prior knowledge on the joint distribution of visual features and labels.
Due to its importance, this is a very active subject of research \cite{lavrenko-2003,monay-2004,carneiro-2007,mei-2008,zhang-2010,dzhang-2012,2pknn-2012,gong2013deep}. Previous work typically use images and associated labels to build classifiers and then assign relevant labels to novel images.
The early works usually rely on images labeled by domain experts \cite{duygulu-2002,monay-2004,carneiro-2007,guillaumin-2009,tousch-2012}, while recently several approaches use weak labels such as user-generated tags in social networks \cite{mcauley-2012,johnson-2015,arxiv2015-li} or query terms in search engines \cite{wang-2008,feifei-2010}.

Despite the source of the labeling, non-parametric models which rely on a nearest-neighbor based voting scheme have received a lot of attention for automatic image annotation \cite{makadia-2008,guillaumin-2009,xli-2009,znaidia-2013,ballan-2015}.
The main reason is that these methods have the ability to adapt to complex patterns as more training data become available. To annotate a new image, they apply a common strategy: first, they retrieve similar images in the training set, and second, they rank labels according to their frequency in the retrieval set.
Automatic image annotation is thus achieved by transferring the most frequent labels in the neighborhood to the test image.
This is essentially a lazy learning paradigm in which the image-to-label association is delayed at test time.
In contrast, discriminative models such as support vector machines \cite{xqi-2007,grangier-2008,sahbi-2010,svmvt-2013} or fully supervised end-to-end deep networks \cite{gong2013deep}, require to define in advance the vocabulary of labels.
This is particularly problematic in a large-scale scenario, such as images on social networks, in which you may have thousands of labels that may also change or increase over time.

Several issues may arise in a nearest-neighbor approach.
The set of retrieved images may contain many incorrect labels, mostly because of the so-called \emph{semantic gap} \cite{cbir-2000}.
This happens because visual features may not be powerful enough in abstracting the visual content of the image.
Thus the proposed algorithms tend to retrieve just the images whose features are very close in the visual space, but the semantic content is not well preserved.
Researchers tried to cope with this issue by improving visual features. To this end, the most significant improvement has been the shift from handcrafting features to end-to-end feature learning, leading to current state-of-the art convolutional neural network representations \cite{krizhevsky2012imagenet,simonyan2014very,ilsvrc}.
Nearest neighbors methods may also suffer when images are not paired with enough label information, leading to a poor statistical quality of the retrieved neighborhood.
This is mostly due to the fact that label frequencies are usually unbalanced.
Modern methods address this issue by introducing label penalties and metric learning \cite{guillaumin-2009,xli-2009,2pknn-2012}.

\begin{figure}
\centering
\includegraphics[width=1\columnwidth]{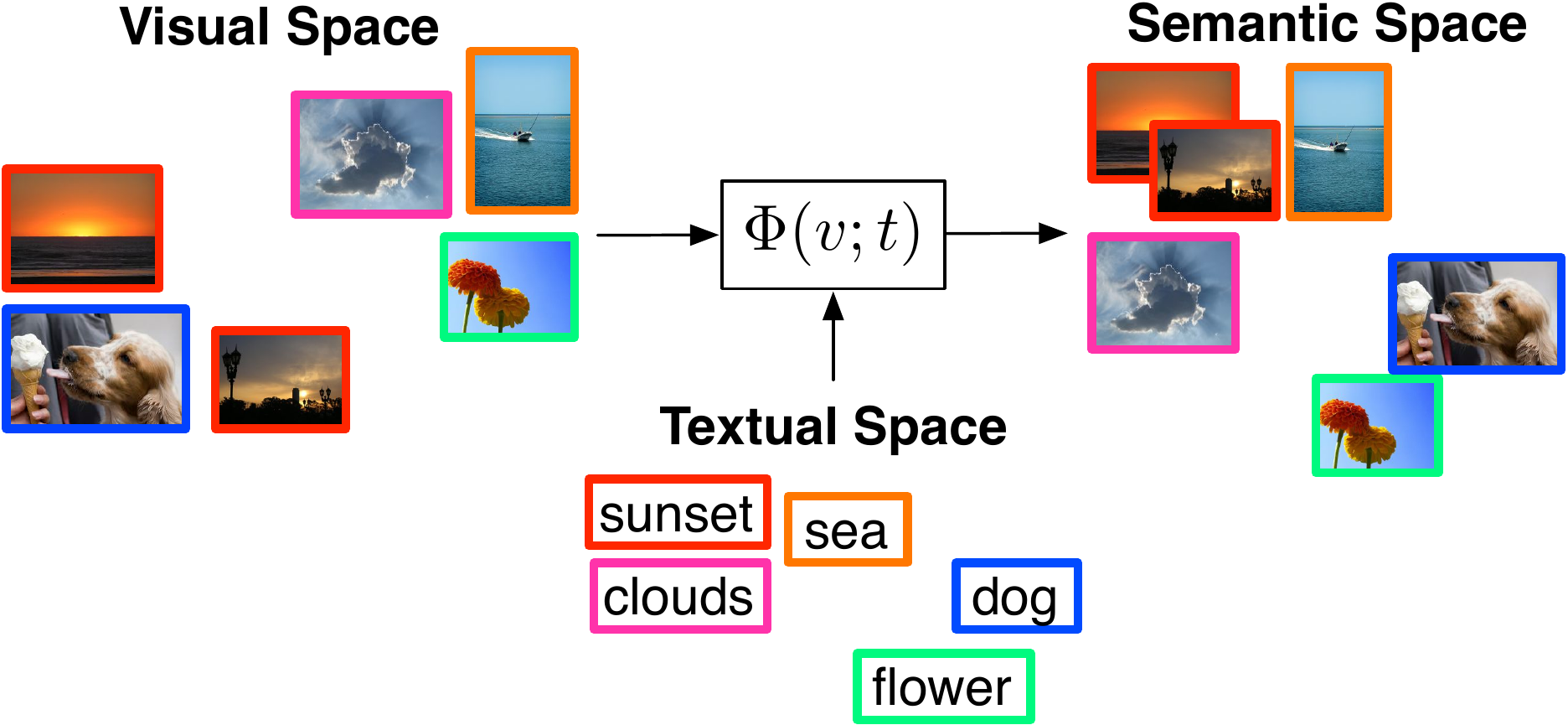}
\caption{Labels associated to the images can be used to re-arrange the visual features and induce the semantics not caught by the original features. For instance, the sunset images with the red border should be closer to images of clouds and sea, according to the text space. A projection $\Phi(v; t)$ is learned to satisfy correlations in visual and textual space.}
\label{fig:intro}
\end{figure}

The image representation can be improved also by shifting to a completely different perspective, namely moving towards a multimodal representation.
A way of bridging the semantic gap might be by designing representations that account not just for the image pixels, but also for its textual representation.
Here we follow this approach by constructing a framework in which the correlation between visual features and labels is maximized. To this end, we present an automatic image annotation approach that relies on Kernel Canonical Correlation Analysis (KCCA) \cite{hardoon-2004}.
Our approach strives to create a semantic embedding through the connection of visual and textual modalities. This embedding lives in a latent space that we refer to as \emph{semantic space}.
Images are mapped to this space by jointly considering the visual similarity between images in the original visual space, and label similarities.
The projected images are then used to annotate new images by using a nearest-neighbor technique or other standard classifiers. 
Figure \ref{fig:intro} illustrates our pipeline.
The main take-home message is that, as illustrated in the figure, the neighborhood of each image will contain more images associated with the same label (e.g. \dquote{sunset}) in the semantic space than in the original visual space (see for example the images with the red border).

\subsection{Main Contributions}

(1) The key contribution of our work is to improve image representations using a simple multimodal embedding based on KCCA. This approach has several advantages over parametric supervised learning. First, by combining a visual and textual view of the data, we reduce the semantic gap. Thus we can obtain higher similarities for images which are also semantically similar, according to their textual representation.
Second, we are free from predetermining the vocabulary of labels. This makes the approach well suited for nearest neighbor methods, which for the specific task of image annotation are more robust to label noise.
A slight disadvantage of our method is its inherent batch nature. Although, as shown in our experimental results, learning the semantic projection is also possible on a subset of the training data.

(2) Previous works that learn multimodal representations from language and imagery exist \cite{srivastava2012dbm}, including prior uses of CCA and KCCA \cite{hardoon-2004,rasiwasia-2010,hwang-2012,gong-2013}.
However, we are the first to propose a framework that combines the two modalities into a joint semantic space which is better exploitable by state-of-the-art nearest neighbor models. Interestingly enough, in our framework the textual information is only needed at training time, thus allowing to predict labels also for unlabeled images.

(3) We provide extensive experimental validations. Our approach is tested on medium and large scale datasets, i.e. IAPR-TC12 \cite{iaprtc12}, ESP-GAME \cite{espgame}, MIRFlickr-25k \cite{mirflickr} and NUS-WIDE \cite{nuswide}.
We show that our framework is able to leverage recently developed CNN features in order to improve the performance even further. Additionally, we introduce a tag denoising step that allows KCCA to effectively learn the semantic projections also from user-generated tags, which are available at no cost in a social media scenario. The scalability of the method is also validated with subsampling experiments.
\smallskip

This paper builds on our previous contribution on cross-modal image representations \cite{ballan-2014} and improves in many ways.
We report new experimental evaluations covering the large dataset NUS-WIDE. Validate our pipeline with modern convolutional neural network based features. Extend our original approach with a new text filtering method that allows the semantic space to be computed from noisy and sparse tags, such as that from social media. Report new insights on several key aspects such as performance and scalability of our approach when subsampling the training set.

\section{Related Work}\label{sec:related}
\subsection{Automatic Image Annotation: Ideas and Main Trends}
Automatic image annotation is a long standing area of research in computer vision, multimedia and information retrieval \cite{arxiv2015-li}.
Early works often used mixture models to define a joint distribution over image features and labels \cite{lavrenko-2003,feng-2004,carneiro-2007}. In these models, training images are used as non-parametric density estimators over the co-occurrence of labels and images.
Other popular probabilistic methods employed topic models, such as pLSA or LDA, to represent the joint distribution of visual and textual features \cite{barnard-2003,monay-2004,yxiang-2009}. They are generative models, thus they maximize the generative data likelihood. 
They are usually expensive or require simplifying assumptions that can be suboptimal for predictive performance.
Discriminative models such as support vector machines (SVM) and logistic regression have also been used extensively \cite{grangier-2008,sahbi-2010,svmvt-2013,izadinia2015deep}.
In these works, each label is considered separately and a specific model is trained on a per-label basis. In testing, they are used to predict whether a new image should be labeled with the corresponding label. While they are very effective, a major drawback is that they require to define in advance the vocabulary of labels. Thus, these approaches do not handle well large-scale scenarios in which you may have thousands of labels and the vocabulary may shift over time.

Despite their simplicity, a class of approaches that has gained a lot of attention is that of nearest-neighbor based methods \cite{makadia-2008,guillaumin-2009,2pknn-2012,ballan-2015}. Their underlying intuition is that similar images are likely to share common labels.
Many of these methods start by retrieving a set of visually similar images and then they implement a label transfer procedure to propagate the most common training labels to the test image.
The most recent works usually implement also a refinement procedure, such as metric learning \cite{guillaumin-2009,2pknn-2012} or graph learning \cite{jliu-2009,jtang-2011,zhu-2014,fsu-2015}, in order to differently weight rare and common labels or to capture the semantic correlation between labels.
They are usually computationally intensive and do not model the intermodal correlation between visual features and labels. In contrast, we introduce a framework in which textual and visual data are mapped to a common semantic space in which labels can be transferred more effectively.

\subsection{Towards More Powerful Visual Representations}
The most recent breakthrough in computer vision came from end-to-end feature learning through convolutional neural networks.
In their seminal paper, Krizhevsky \etal \cite{krizhevsky2012imagenet} demonstrated unprecedented improvement in large-scale image classification on ImageNet \cite{deng2009imagenet} using CNNs. These networks are composed of a hierarchy of layers, alternating convolutions and subsampling. They require high quality supervision with minimal noise in labeling. Since then, many researchers have applied deep learning to other visual recognition tasks such as object detection and image parsing \cite{girshick2014rich}. Deeper architectures have been recently proposed, showing further gain in image classification accuracy (e.g. \cite{simonyan2014very}).

Another interesting property of these architectures is that they have the ability to learn representations that can be transferred and used in many other tasks, such as attribute prediction and image retrieval \cite{razavian2014cnn}.
Convolutional neural networks (CNNs) have been also recently applied to automatic image annotation \cite{gong2013deep}, showing significant improvement in terms of precision and recall.
On top of these powerful features, a number of recent works have used more advanced encoding schemes in order to improve feature generalization.
For instance, VLAD encoding is applied in \cite{gong2014multi} to pool multi-scale CNN features computed over different windows, while Fisher Vector encoding applied to dense multi-scale CNN activations is used in \cite{yoo-cvprw-2015}.
This has been also improved in \cite{uricchio-2015} by applying Fisher Vector to sparse boxes, selected by objectness or random selection.
However, all these approaches only focus on the visual modality.

\subsection{Cross-media and Multimodal Representations}
A number of approaches have been developed for learning multimodal representations from images and labels \cite{lavrenko-2003,carneiro-2007,mcauley-2012,srivastava2012dbm,frome2013devise,guillaumin-2010}.
In particular, we highlight that previous use of CCA and its variants exists, particularly for the task of cross-modal image retrieval \cite{hardoon-2004,rasiwasia-2010,hwang-2012,gong-2013,habibian2015discovering,ccir} and multi-view learning~\cite{twocca,mvisl}.
This class of methods is often used to learn multi-view embeddings in a unimodal setting.
For example, Yang~\etal \cite{twocca} use CCA to learn a common representation from two views in the image space.
A more general approach is presented in~\cite{mvisl} where a latent representation of samples is learned from multiple views. Their framework can be applied also to combine visual features or imagery captured in different conditions.

Hardoon \etal were the first to apply KCCA to image retrieval with textual query \cite{hardoon-2004}. 
Successively, Rasiwasia \etal \cite{rasiwasia-2010} proposed to employ LDA and CCA to perform cross-modal retrieval on text and images obtaining improved results on single modalities. 
In \cite{hwang-2012}, a method to learn importance of textual object is proposed. They show that features such as word frequency, relative and absolute label rank are helpful to evaluate importance of textual information.
Multi-modal learning has been applied to improve ranking in image retrieval fusing visual features and click features in~\cite{ccir}.
A three-way CCA is proposed in \cite{gong-2013} to address the limited expressiveness of CCA. They show that adding a third view representing categories or clustered labels can improve retrieval performance.
Murthy \etal~\cite{murthy-2015} propose to combine CNN features and word embeddings using CCA, but their approach is only tested on small scale datasets using expert labels.
Embeddings carry many advantages, nonetheless learning such coupled representation may be extremely computationally expensive. Recently, there have been some attempts at making such approaches scalable~\cite{frome2013devise, weston2010large}. These on-line methods have usually low memory footprint, and scale very well to large dataset. Nonetheless, they are not designed to tackle multi-label image annotation and they are not able to learn from noisy examples such as tags extracted from social media.

Differently from prior work, we tackle the specific problem of multi-label image annotation.
For this task, only visual features are available at test time.
Thus, our approach exploits labels only at training time.
To this end, we learn a re-organization of the visual space to that of a semantic space where images that share similar labels are closer.
Moreover, when combined to a nearest-neighbors scheme, our approach can predict labels that were not available at training time, when the projections have been learned.

\begin{figure*}[!t]
\centering
\includegraphics[width=0.8\textwidth]{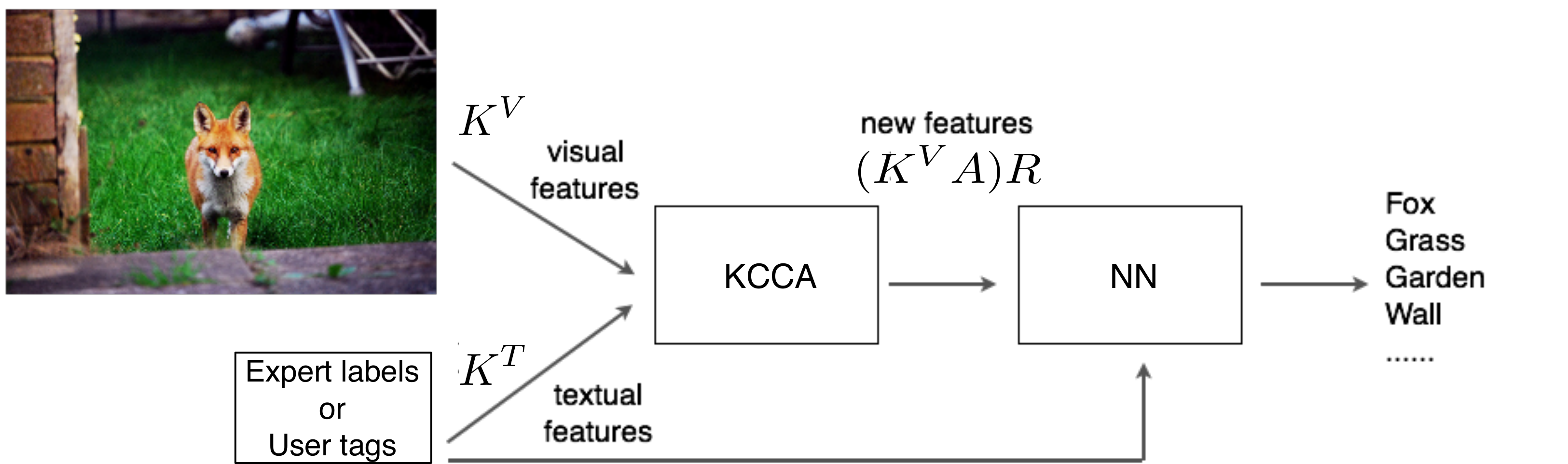}
\caption{Overview of our approach. Image and textual features are projected onto a common \emph{semantic space} in which nearest-neighbor voting is used to perform label transfer.}
\label{fig:approach}
\end{figure*}

\section{Approach}\label{sec:approach}

Our key intuition is that the semantic gap of visual features can be reduced by constructing a semantic space that comprises the fusion of visual and textual information.
To this end, we learn a transformation that embeds textual and visual features into a common multimodal representation.
The transformation is learned using KCCA \cite{hardoon-2004}.
This algorithm strives to provide a common representation for two views of the same data.
Similarly to \cite{hardoon-2004,hwang-2012}, we use KCCA to connect visual and textual modalities into a common \emph{semantic space}, but differently from them, which focus on cross-modal retrieval, our framework is designed to effectively tackle the particular problem of image annotation.
Moreover, we are able to construct the semantic space even exploiting noisy labels, such as the user tags.
Advanced nearest neighbors methods are then used to perform label transfer. An overview of the approach is shown in Fig.~\ref{fig:approach}.

Throughout the paper, we use the term \emph{labels} when we refer to generic textual information. We explicitly use the terminology \emph{expert labels} and \emph{user tags} when we refer only to the expert provided labels or the tags provided by users in social network, respectively.
We now proceed in detailing the visual and textual representation, how KCCA is used to build the semantic space, and finally we describe our label transfer procedures.

\subsection{Visual Features}
We use a deep convolutional neural network pre-trained on ImageNet~\cite{deng2009imagenet} with the VGG-Net architecture presented in \cite{simonyan2014very} (using 16 layers)\footnote{In our preliminary experiments we found that this configuration gives the best results on all our datasets, although other networks gave similar results.}.
We use the activations of the last fully connected layer as image features.
Such representation proved to be good for several visual recognition and classification tasks \cite{razavian2014cnn,girshick2014rich}.

Given an image $I_i$, we first warp it to $224\times224$ in order to fit the network architecture and subtract the training images mean. 
We use this normalized image to extract the activations of the first fully connected layer.
Let $\phi^V({I_i})$ be the extracted feature of $I_i$.
We use the ArcCosine kernel:
\begin{equation} \label{eq:kernel_visual}
K^V_n(\phi^V(I_i), \phi^V(I_j)) = \frac{1}{\pi} ||\phi^V(I_i)||^n ||\phi^V(I_j)||^n J_n(\theta)
\end{equation}
where $J_n$ is defined according to the selected order of the kernel.
Following \cite{cho2009kernel}, we set $n=2$ which gives us:
\begin{equation}
J_2(\theta) = 3 \sin \theta \cos \theta + (\pi - \theta)(1 + 2 \cos^2 \theta)
\end{equation} 
where $\theta$ is the angle between the inputs $\phi^V(I_i)$, $\phi^V(I_j)$. This kernel provides a representation that is better suited to neural networks activations and gives better results.
We also tried other kernels such as linear and radial basis function, obtaining a slightly inferior performance ($\sim$1\%).

\subsection{Textual Features}
Depending on how labels are generated, \emph{i.e.} expert labels or user-generated tags, we should use different approaches. While expert labels can be trusted, user-generated tags are noisy and require a more robust representation.

\begin{figure*}
\centering
\subfigure[CNN Features]{\label{fig:kcca:a}\includegraphics[width=0.5\columnwidth]{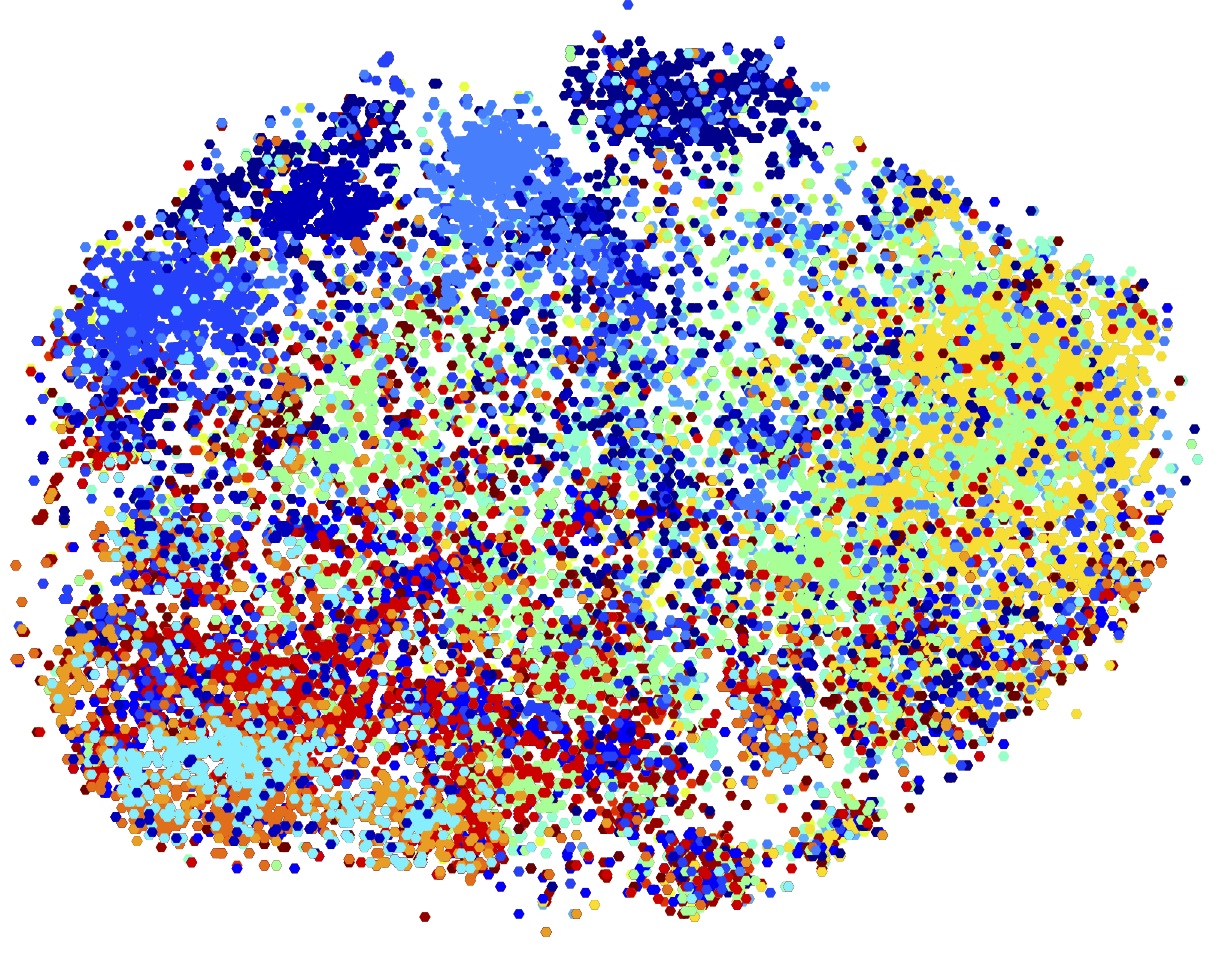}}
\hspace{2pt}
\subfigure[KCCA + Expert Labels]{\label{fig:kcca:b}\includegraphics[width=0.5\columnwidth]{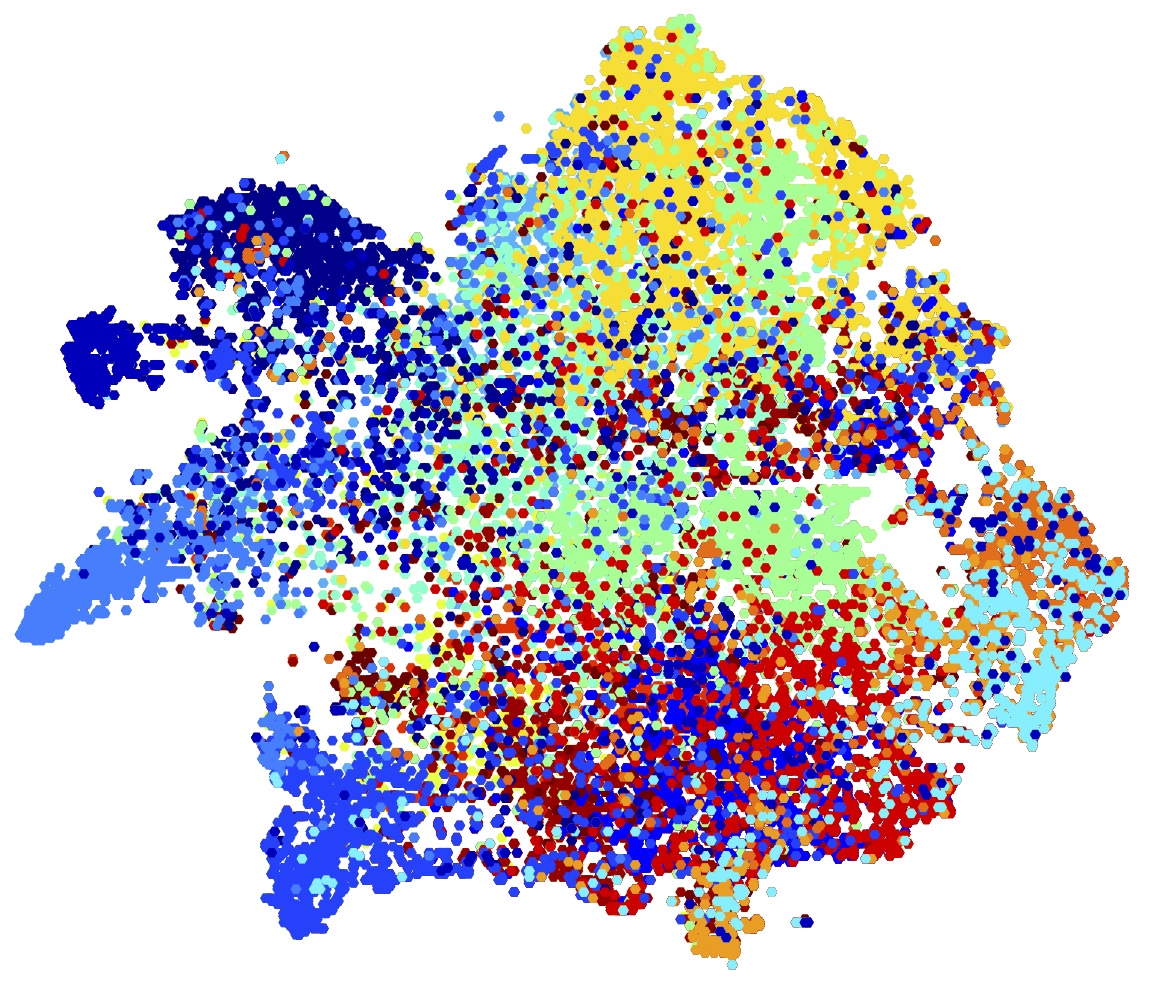}}
\subfigure[KCCA + User Tags]{\label{fig:kcca:c}\includegraphics[width=0.5\columnwidth]{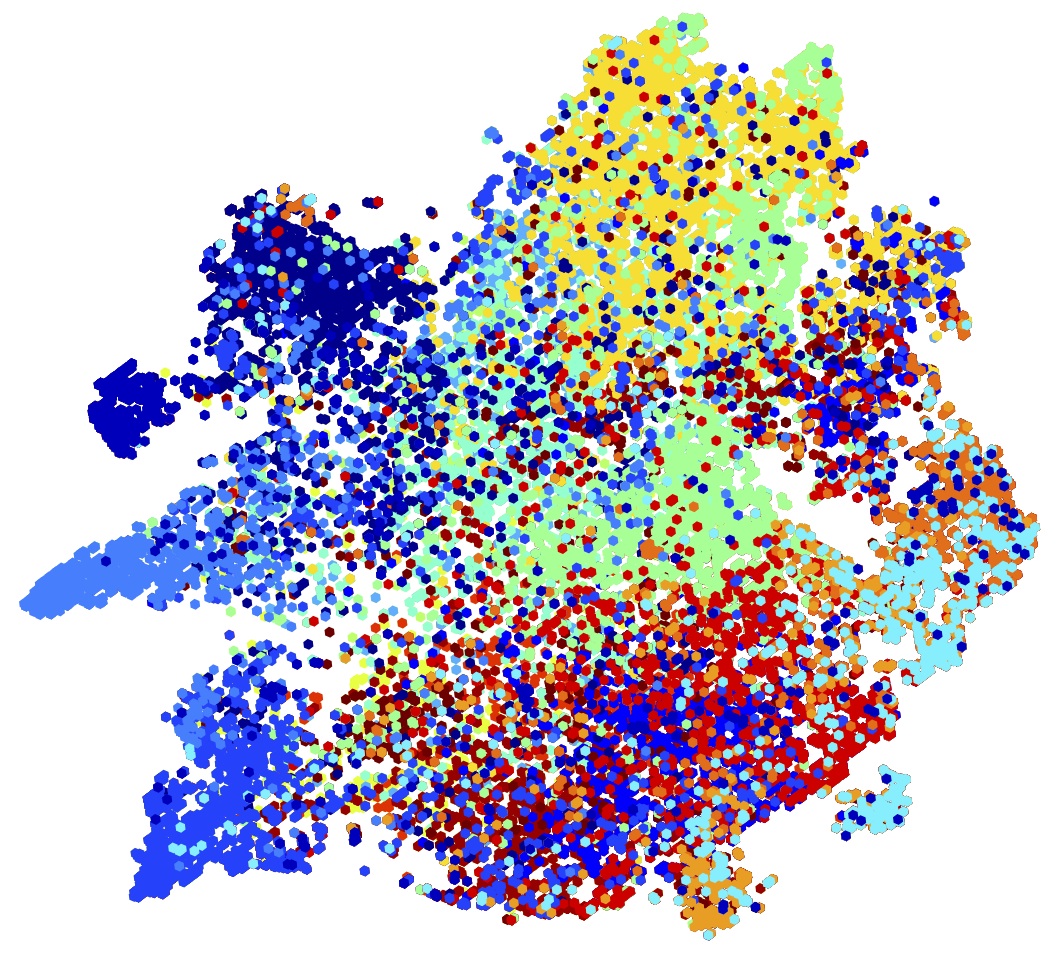}}
\caption{t-SNE visualization of images on MIRFlickr-25K with different features. Each color corresponds to a different label.}
\label{fig:kcca}
\end{figure*}

\subsubsection{Expert Labels}\label{sec:expertlabels}
For expert labels, we use simple binary indicator vectors as textual features.
Let $D$ be the vocabulary size, \emph{i.e.} the number of labels used for annotation. 
We map each label set of a particular image $I_i$ to a $D$-dimensional feature vector $\phi^T(I_i)=[w^i_1,\cdots,w^i_D]$, where $w_k$ is $0$ or $1$ if that image has been annotated with the corresponding $k$-th label $l_k$.
This results in a highly sparse representation.
Then we use a linear kernel which corresponds to counting the number of labels in common between two images:
\begin{equation}
K^T(\phi^T(I_i), \phi^T(I_j)) = \sum_{k=1}^D w^i_k w^j_k.
\end{equation}

The basic idea is that we are considering the co-occurrences of labels in order to measure the similarity between two images.
Nonetheless, this representation models each label independently from the others. It has been shown in previous works that exploiting semantic relations by weighting each label differently can improve performance \cite{johnson-2015,hu-2015}. Therefore, we explore two textual kernels that consider semantic relations between labels: an ontology-based textual kernel with bag-of-words \cite{shawe2004kernel} and one that exploits the more recent continuous word vector representation \cite{mikolov-2013}.
For the bag-of-words semantic kernel, the idea is to weight each label in a linear kernel by using a similarity matrix $S \in \mathbb{R}^{D \times D}$ as:
\begin{equation}
	K^T(\phi^T(I_i), \phi^T(I_j)) = \phi^T(I_i) S \phi^T(I_j)^\intercal.
\end{equation} 
We set the elements of $S$ as the Lin similarity \cite{lin1998information} between each label, using WordNet.
This measure has been used successfully in several works to suggest similar labels (see \cite{arxiv2015-li}).
Regarding the continuous word vector kernel, Mikolov \etal \cite{mikolov-2013} recently showed that it is possible to learn a word representation from a large scale corpus in an unsupervised way. The learned word vector features were proved to model semantics in form of regularities in several applications \cite{frome2013devise,murthy-2015}.
Given the learned representation of a label $w_k$ as $\zeta(w_k) \in \mathbb{R}^P$, we represent the set of labels of an image $I_i$ using average pooling 
\begin{equation}
	\phi^T(I_i) = \frac{1}{N} \sum_k^D w^i_k \cdot \zeta(l^i_k).
\end{equation} 
Finally, we apply a linear kernel on such representation:
\begin{equation}
	K^T(\phi^T(I_i), \phi^T(I_j)) = \phi^T(I_i) \phi^T(I_j)^\intercal.
\end{equation} 
We compare the performance obtained with these three textual representations in Sect. \ref{sec:results_txtkernels}.

\subsubsection{Denoising User-generated Tags}
For user-generated tags, we should first reduce the labeling noise. To this end, we perform a ``pre-propagation'' step based on visual similarity. 
The purpose of this tag denoising step is two-fold: first, we need to improve the quality of tags of each training image in order to learn a proper embedding; second, we need to cope with the sparsity of user tags. For the first issue, our assumption is that by gathering a neighborhood of visually similar images the more frequent tags will fade out noisy tags in favor of content related ones. Regarding the sparsity issue, images usually are labeled with few tags and in extreme cases they can have no tags at all. For this reason, the visual information is the most reliable information we can exploit.

Thus, we shall obtain a cleaner tag feature-vector $\hat{\phi}^T(I_i) = [\hat{w}_{i,1},\cdots,\hat{w}_{i,D}]$ and then compute the textual kernel $K^T$. We start from the representation $\phi^T(I_i)=[w^i_1,\cdots,w^i_D]$, where $w_k$ is $0$ or $1$ if the image $I_i$ has been annotated with the corresponding tag $t_k$.
For each image $I_i$ we consider the $R$=100 most similar images, according to the visual kernel $K^V$ (the same pre-computed in Eq. \ref{eq:kernel_visual}), and compute the new tag vector:
\begin{equation}
\hat{\phi}^T(I_i) = \frac{\sum_{k=1}^R x_k \phi^T(I_k)}{\sum_{k=1}^R x_k}
\end{equation}
where $x_k=\exp(-\frac{||\phi^V(I_i) - \phi^V(I_k)||^2}{\sigma})$ is an exponentially decreasing weight computed from image similarities. We set $\sigma$ to the mean of the distances.
This improved tag vector can be seen as an approximation of the probability mass function of tags among its nearest neighbor images. We use the exp-$\chi^2$ kernel:
\begin{equation}
K^T(\hat{\phi}^T(I_i),\hat{\phi}^T(I_j))= \exp\left(-\frac{1}{2C}\sum_{k=1}^{D}{\frac{(\hat{w}_{i,k}-\hat{w}_{j,k})^2}{(\hat{w}_{i,k}+\hat{w}_{j,k})}}\right)
\end{equation}
where $C$ is set to the mean of the $\chi^2$ distances.
We demonstrate in section \ref{sec:results_tags} that this pre-propagation step is essential to learn the semantic embedding properly, as clearly shown by the results reported in Table \ref{tab:denoising_mirflickr}.

\subsection{Kernel Canonical Correlation Analysis}\label{sec:kcca}
Given two views of the data, such as the ones provided by visual and textual features, we can construct a common multimodal representation. We first briefly describe CCA and then move to explain the extended KCCA algorithm.
CCA seeks to utilize data consisting of paired views to simultaneously find projections from each feature space so that the correlation between the projected representations is maximized.

\iftoggle{single}{
\newcommand{\figwidth}{0.22}
}{
\newcommand{\figwidth}{0.31}
}

More formally, given $N$ training pairs of visual and textual features $\{(\phi^V(I_1),\phi^T(I_1)),\dots,(\phi^V(I_N),\phi^T(I_N))\}$, 
the goal is to simultaneously find directions $z_V^*$ and $z_T^*$ that maximize the correlation of the projections of $\phi^V$ onto $z_V^*$ and $\phi^T$ onto $z_T^*$. This is expressed as:
\begin{align}
z_V^*,z_T^* = \arg\max_{z_V,z_T}\frac{\mathrm{E}[ \langle \phi^V,z_V\rangle \langle \phi^T,z_T\rangle ]}{\sqrt{\mathrm{E}[ \langle \phi^V,z_V\rangle^2] \mathrm{E}[\langle \phi^T,z_T\rangle^2 ]}} \nonumber \\=
\arg\max_{z_V,z_T}\frac{z_V^\intercal C_{vt} z_T}{\sqrt{z_V^\intercal C_{vv} z_V z_T^\intercal C_{tt} z_T}}
\end{align}
where $ \mathrm{E} [\cdot] $ denotes the empirical expectation, while $C_{vv}$ and $C_{tt}$ respectively denote the auto-covariance matrices for $\phi^V$ and $\phi^T$, and $C_{vt}$ denotes the between-sets covariance matrix. 

The CCA algorithm can only model linear relationships. As a result, KCCA has been introduced to allow projecting the data into a higher-dimensional feature space by using the kernel trick \cite{hardoon-2004}.
Thus, the problem is now to search for solutions of $z_V^*$ and $z_T^*$ that lie in the span of the $N$ training instances $\phi^V(I_i)$ and $\phi^T(I_i)$:
\begin{align}
z_V^* = \sum_{i=1}^N\alpha_{i}\phi^V(I_i), \qquad 
z_T^* = \sum_{i=1}^N\beta_{i}\phi^T(I_i).
\end{align}
The objective of KCCA is to identify the weights $\alpha,\beta \in \mathbb{R}^N$ that maximize:
\begin{equation}
\alpha^*,\beta^* = \arg\max_{\alpha,\beta}\frac{\alpha^\intercal K^V K^T \beta}{\sqrt{\alpha^\intercal (K^V)^2 \alpha \beta^\intercal (K^T)^2 \beta}}
\end{equation}
where $K^V$ and $K^T$ denote the $N \times N$ kernel matrices over a sample of $N$ pairs.
As shown by Hardoon \etal \cite{hardoon-2004}, learning should be regularized in order to avoid trivial solutions. Hence, we penalize the norms of the projection vectors and obtain the generalized eigenvalue problem:
\begin{equation} \label{eq:kcca_regularization}
(K^V + \kappa I)^{-1}K^T(K^T + \kappa I)^{-1}K^V\alpha = \lambda^2 \alpha
\end{equation}
where $\kappa \in [0, 1]$. The top $M$ eigenvectors of this problem yield bases $A=\left[\alpha_1\dots\alpha_M\right]$ and $B=\left[\beta_1\dots\beta_M\right]$ that we use to compute the semantic projections of training and test kernels. For each pair $(\alpha_j, \beta_j)$ of the given bases, the corresponding eigenvalue $r_j$ measures the correlation between projected input pairs. 
Higher $r_j$ is associated with higher correlation, thus it is convenient to weight more the dimensions of higher energy.
According to this principle, we obtain the final features as:
\begin{equation} \label{eq:semantic_space}
\psi(I) = (K^V A) R
\end{equation}
where $R = \text{diag}([r_1, \ldots, r_M])$.
Note that $\psi$ has no dependency on the textual space. Thus, projecting new test images requires only their visual features $\Phi^V$, making our approach suitable for automatic image annotation.

In Figure \ref{fig:kcca} we show t-SNE embeddings \cite{jmlr2014-maaten} of the CNN features and their projection into the semantic space.
These plots qualitatively show that KCCA improves the separation of the classes, both in case of expert labels and user-generated tags.
This leads to a more accurate manifold reconstruction and, as our experiments will confirm, a significant improvement in performance.

\subsection{Label Transfer}\label{sec:kcca-annot}
\begin{figure}
\centering     
\subfigure[Baseline]{\label{fig:neighbors:a}\includegraphics[width=0.46\columnwidth]{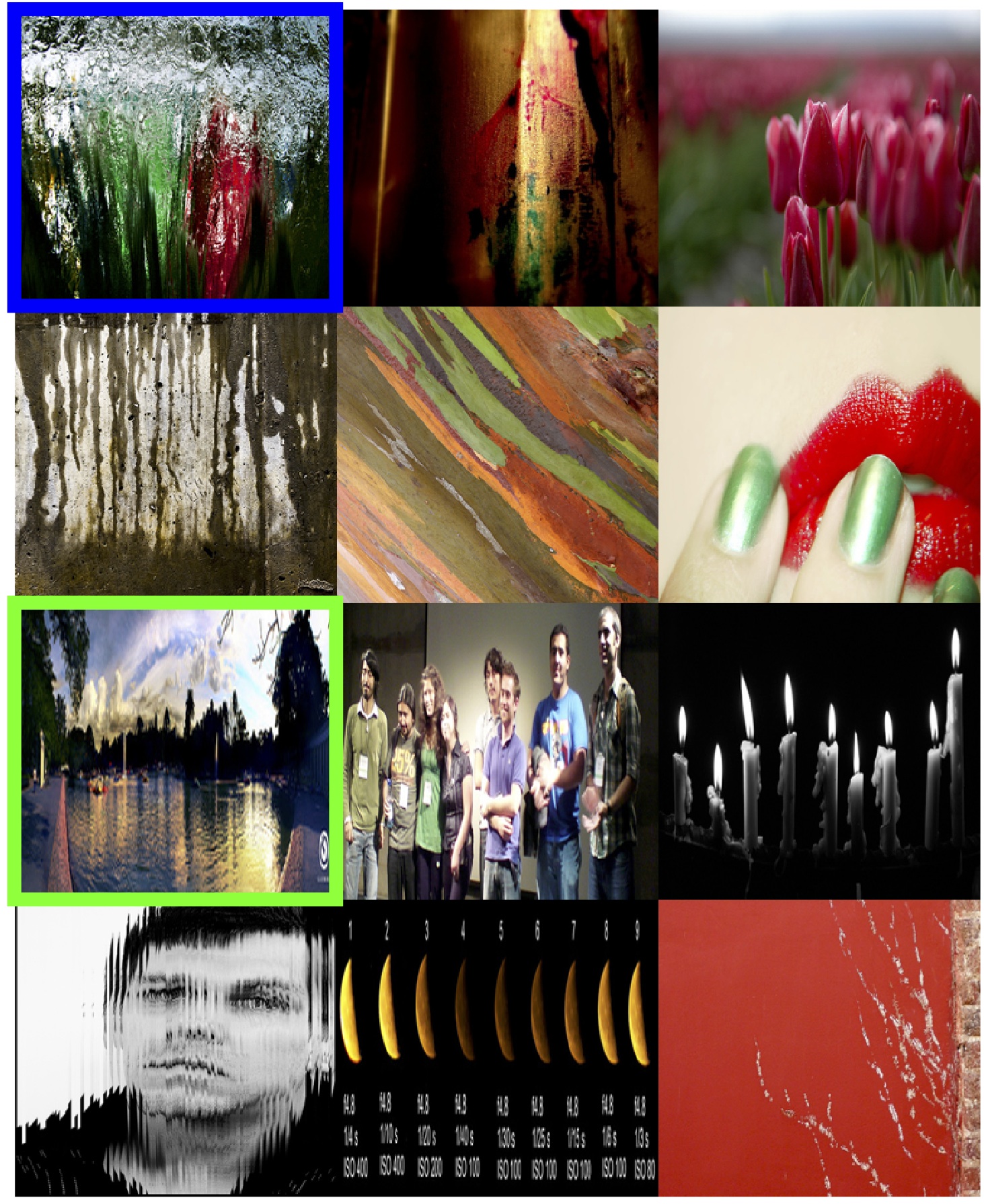} }
\hspace{2pt}
\subfigure[Our Method]{\label{fig:neighbors:b}\includegraphics[width=0.46\columnwidth]{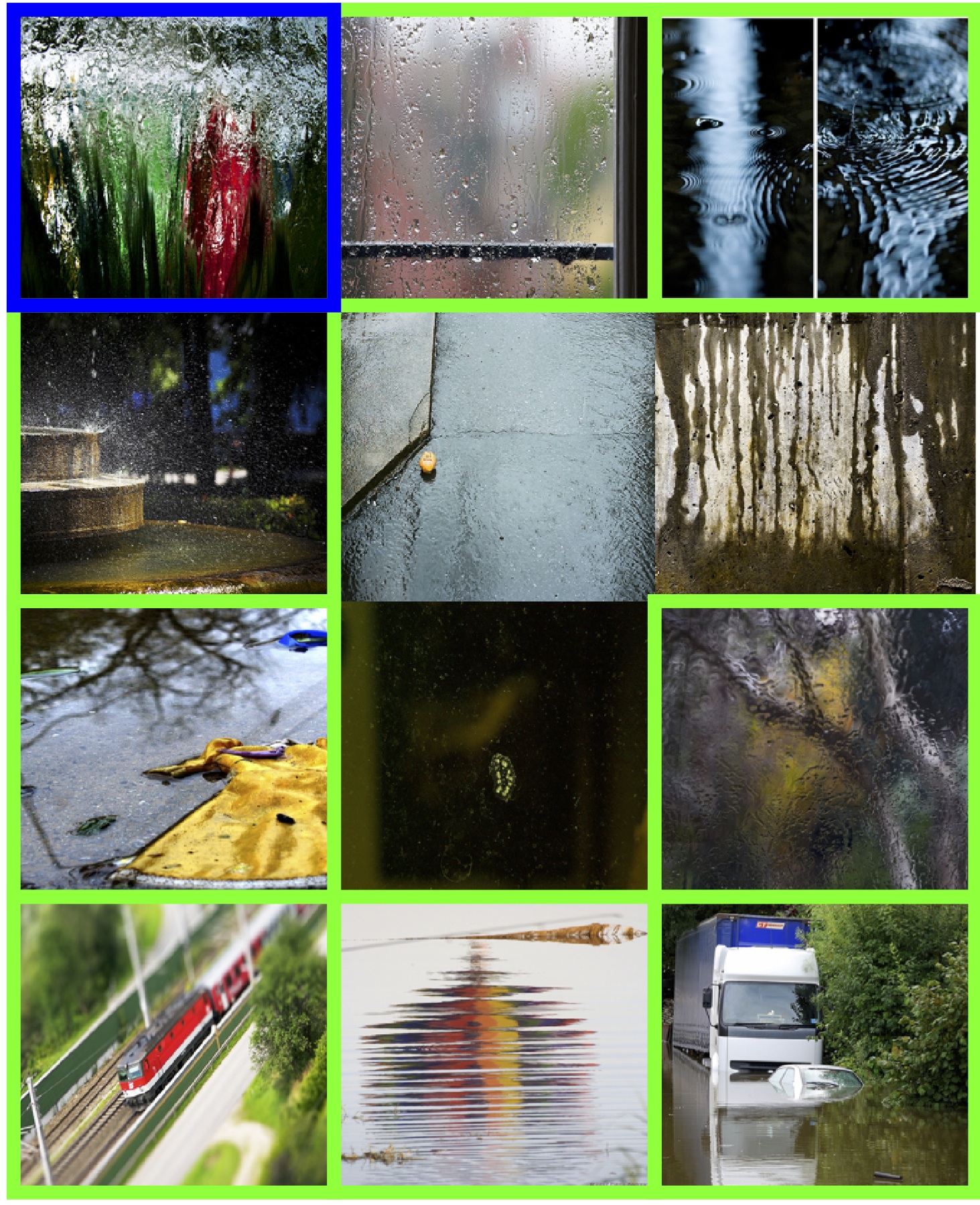}}

\caption{Nearest neighbors found with baseline representation (a) and with our proposed method (b) for a water image (first highlighted in blue in both figures) from the MIRFlickr-25K dataset. Training images with ground truth label \textit{water} are highlighted with a green border. Nearest neighbors are sorted by decreasing similarity.}
\label{fig:neighbors}
\end{figure}

The constructed semantic space assures that similar images, in visual space or in textual space, have also similar features. This property is especially useful for the class of nearest-neighbor methods, since they rely on the intuition that similar images share common labels. 
We show examples of this property in Figure \ref{fig:neighbors}. 
We compare the neighbors retrieved for the same query using the baseline visual features and the semantic space features from our method.
The query, depicted in a blue box, is an image of water where green and red lights produce a fascinating visual effect.
The other images are the most similar images retrieved by one of the two settings. We put a box in green on images that have the correct label ``water'' associated.
We see that neighbors retrieved in the baseline space share some visual similarity: they mostly have green and red colors, some line or dotted patterns that mimic the query image. However only one image is really about water. Our method, instead, successfully retrieves 8 of 11 images with the label water, even if they are quite dissimilar in the visual space.
Indeed, it is impossible with the images in Figure \ref{fig:neighbors:a} to obtain a meaningful neighborhood since the correct label ``water'' is not frequent enough to be relevant in the final labels rank. 

\begin{figure}
\centering
\includegraphics[width=\columnwidth]{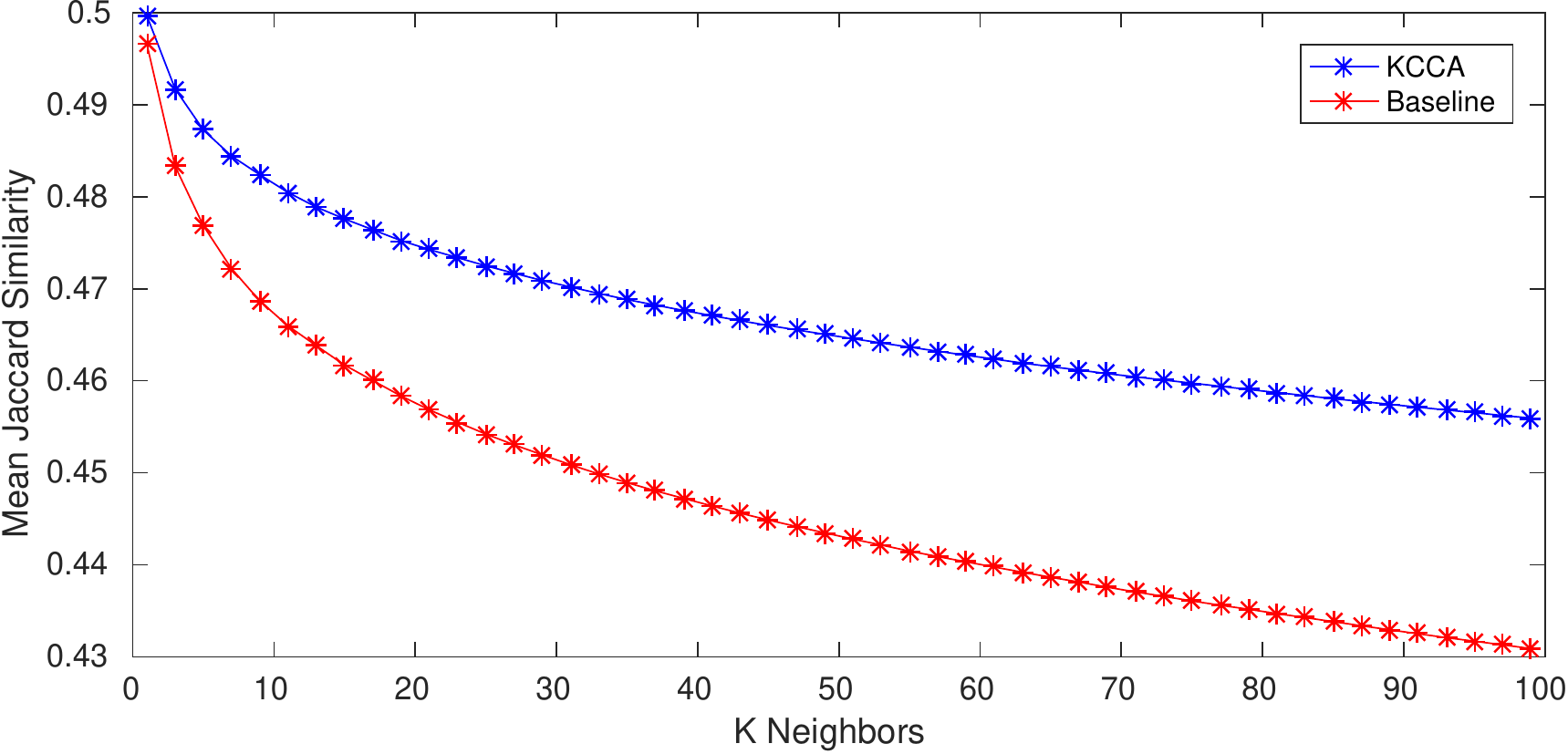}\vspace{-5pt}
\caption{Mean Jaccard similarity between label sets of a test image and the label sets of images in the neighborhood build using visual and KCCA features varying the neighborhood size.}\label{fig:jaccard}
\end{figure}

A quantitative characterization of this behavior can be seen comparing the sets of labels of images in the neighborhood of a test image with the correct labels of the image itself. We run an experiment on NUS-WIDE, measuring this similarity using Jaccard distance.
Specifically, for each image $\hat{x}$ of the test set, we retrieve the $K$ most similar images $\{x_1, x_2, \ldots, x_K\}$ using the visual features and then compute the mean Jaccard similarity between their sets of labels as:$$\frac{1}{K} \sum_{i=1}^K J(\hat{\mathcal{Y}}, \mathcal{Y}_i) = \frac{1}{K} \sum_{i=1}^K \frac{|\hat{\mathcal{Y}} \cap \mathcal{Y}_i|}{|\hat{\mathcal{Y}}| + |\mathcal{Y}_i| - |\hat{\mathcal{Y}} \cap \mathcal{Y}_i|},$$ where $\hat{\mathcal{Y}}$ and $\mathcal{Y}_i$ are, respectively, the set of labels of $\hat x $ and $x_i$. We compute this measure for each test image and average them in a final similarity index as reported in Fig.~\ref{fig:jaccard}.

The higher Jaccard similarity yielded by KCCA features  with respect to baseline visual features, shows that the neighbors retrieved using KCCA have a label distribution which is closer to the one of the query.

Following this key idea, we have used four nearest-neighbor voting algorithms in our semantic space in order to automatically annotate images. 
Nevertheless, we expect that other general class of learning algorithms may take advantage of the semantic space. To this end, we also consider the off-the-shelf SVM classifier. 
Given an image and a vocabulary of labels, each algorithm performs automatic image annotation by applying a particular \emph{relevance function} \cite{arxiv2015-li}, as defined in the following.

\medskip\subsubsection{Nearest-Neighbor Voting}\label{nn-vote}
The most straightforward approach is to project the test image onto the semantic space, and then identify its $K$ nearest-neighbors.
Here we rank the vocabulary labels according to the their frequency in the retrieval set.
Thus, the relevance function is defined as:
\begin{equation}
f_{KNN}(I,t) := k_t
\end{equation}
where $k_t$ is the number of images labeled as $t$ in the neighborhood of $I$. 

\medskip\subsubsection{Tag Relevance}\label{nn-li}
Li \etal \cite{xli-2009} proposed a relevance measure based on the consideration that if several people label visually similar images using the same labels, then these labels are more likely to reflect objective aspects of the visual content.
Following this idea it can be assumed that, given a query image, the more frequently the tag occurs in the neighbor set, the more relevant it might be. However, some frequently occurring labels are unlikely to be relevant to the majority of images. To account for this fact, the proposed tag relevance measurement takes into account both the number of images with tag $t$ in the visual neighborhood of $I$ (namely $k_t$) and in the entire collection:
\begin{equation} 
f_{TagVote}(I, t) := k_t - {K} \frac{n_t}{|\mathcal{S}|}\label{eq:tagvote}
\end{equation}
where $n_t$ is the number of images labeled with $t$ in the entire collection $\mathcal{S}$ and $K$ is the number of neighbors retrieved.

\medskip\subsubsection{TagProp}\label{nn-tagprop}
Guillaumin \etal \cite{guillaumin-2009} proposed an image annotation algorithm in which the main idea is to learn a weighted nearest neighbor model, to automatically find the optimal metric that maximizes the likelihood of a probabilistic model. The method can learn rank-based or distance-based weights:
\begin{equation}
f_{TagProp}(I, t) := \sum_j^K  \pi_j \cdot \mathcal{I}(I_j,t)
\end{equation}
where $K$ is the number of neighbors retrieved,  $\mathcal{I}$ is the indicator function that returns 1 if $I_j$ is labeled with $t$, and 0 otherwise; $\pi_j$ is a learned weight that accounts for the importance of the $j$-th neighbor $I_j$.
In addition the model can be extended with a logistic per-tag model to promote rare labels and suppress the frequent ones.

\medskip\subsubsection{2PKNN}\label{nn-2pknn}
Verma and Jawahar \cite{2pknn-2012} formulated the problem as a probabilistic framework and proposed a two-phase approach: given a test image, a first phase is employed to construct a balanced neighborhood. Then, a second phase uses image distances to perform the actual estimation of the tag relevance. 
Given a test image $I$ and a vocabulary of $D$ labels, the first phase collects a set of neighborhoods $\mathcal{N}(I)$ composed of the nearest $M$ training images annotated with each $t$ in $D$. On the second phase, the balanced neighborhood is used to estimate the tag relevance of $t$ to $I$:
\begin{equation}
f_{2PKNN}(I, t) := \sum_{I_j\in \mathcal{N}(I)} \exp(-d(I, I_j)) \cdot \mathcal{I}(I_j, t)
\end{equation}
where $d(I,I_j)$ is a distance function between image $I$ and $I_j$. Since the distance function is parametrized with a trainable weight for each dimension, the algorithm presented in \cite{2pknn-2012} also performs metric learning similarly to TagProp (we refer to the complete algorithm as to 2PKNN-ML). We only consider the version without metric learning, since our implementation of 2PKNN-ML performs worse than 2PKNN.

\medskip\subsubsection{SVM}\label{svm}
For each label, a binary linear SVM classifier is trained using the L2-regularized least square regression, similarly to \cite{verbeek-2010}. Independently from the source of labels, be it expert labels or user tags, the images with the label are treated as positive samples while the others as negative samples.
To efficiently train our classifier we use stochastic gradient descent (SGD). The relevance function is thus:
\begin{equation} \label{eq:svm-af}
	f_{SVM}(I,t) := b + \langle w_t, \psi(I) \rangle,
\end{equation}
where $w_t$ are the weights learned for label $t$ and $b$ is the intercept.

\section{Experiments}\label{sec:experiments}

\subsection{Datasets}
Automatic image annotation with expert labels has been historically benchmarked with three datasets: Corel5K, ESP-GAME and IAPR-TC12.
We follow previous work but discard Corel5K since it is outdated and not available publicly. Note that these datasets have poor quality images and they lack metadata as well as user tags. 
Thus, we additionally consider two popular datasets collected from Flickr, i.e. MIRFlickr-25k and NUS-WIDE.
Dataset statistics are summarized in Table \ref{tab:datasets}.

\begin{table}[t]
\centering
\caption{Datasets Statistics.}
\label{tab:datasets}
\resizebox{0.45\textwidth}{!}{
\begin{tabular}{cccccc}
\toprule
 &  &  &  & \textbf{Expert} & \textbf{User}\\
\textbf{Dataset}       & \textbf{Images}  & \textbf{Labels} & \textbf{Tags} &  \textbf{Labels} & \textbf{Tags}\\
\midrule
IAPR-TC12     & 19,627  & 291 & - & \checkmark         & -         	 \\
ESP-GAME      & 20,770  & 268 & - & \checkmark         & -         	 \\
MIRFlickr-25k & 25,000  & 18   & 1,386 & \checkmark          & \checkmark     \\
NUS-WIDE      & 269,648 & 81   & 5,018 & \checkmark        & \checkmark     \\
\bottomrule
\end{tabular}
}
\end{table}

\medskip\textbf{ESP-GAME.}~The ESP-GAME dataset \cite{espgame} was built through an online game. Two players, not communicating with each other, describe images through labels and obtain points when they agree on the same terms.
Since the image is the only media the players see, they are pushed to propose visually meaningful labels. Following previous work, we used the same split of \cite{guillaumin-2009} consisting of $18,689$ images for training and $2,081$ for test. There is an average of $4.68$ annotated labels per image out of $268$ total candidates.

\medskip\textbf{IAPR-TC12.}~This dataset was introduced in \cite{iaprtc12} for cross-language information retrieval. It is a collection of $19,627$ images comprised of natural scenes such as sports, people, animals, cities or other contemporary scenes. 
Like previous work, we used the same setting as in \cite{guillaumin-2009}. It consists of $17,665$ training images and $1,962$ testing images. Each image is annotated with an average of $5.7$ labels out of $291$ candidates.

\medskip\textbf{MIRFlickr-25K.}~The MIRFlickr-25K dataset \cite{mirflickr} has been introduced to evaluate keyword-based image retrieval.
It contains $25,000$ images downloaded from Flickr, $12,500$ images for training and the same amount for testing. For each image, the presence of $18$ labels are available as expert labels as well as user tags (we consider the same labels as in \cite{verbeek-2010}).
They are annotated with an average of respectively $2.78$ expert labels and $8.94$ user tags. Note that tags corresponding to the expert labels are very scarce in this dataset.
Beside tag annotations, EXIF information and other metadata such as GPS are available. While the ground-truth labels are exact, the user tags are weak, noisy and overly personalized. Moreover, not all of them are relevant to the image content. We used the same training and test sets as in previous work \cite{verbeek-2010}.

\medskip\textbf{NUS-WIDE.}~The NUS-WIDE dataset \cite{nuswide} is composed of $269,648$ images retrieved from Flickr. Similarly to MIRFlickr, $81$ labels are provided as expert labels as well as user tags.
Images are annotated with an average of $2.40$ expert labels and $8.48$ user tags, respectively.
NUS-WIDE is one of the largest datasets of images collected from social media. The sparsity of labels and user tags is one of the main challenges in exploiting this dataset as a training set. Moreover the distribution of labels is unbalanced with few concepts being present in almost 80\% of the images: \dquote{sky}, \dquote{clouds}, \dquote{person} and \dquote{water}.
Following previous work, we discard images without any expert label \cite{gong2013deep}, leaving us with $209,347$ images that we further split into $\sim$125K for training and $\sim$80K for testing, by using the split provided by the authors of the dataset. 

\begin{table*}[t!]
\centering
\caption{Results of our method compared to the state of the art on IAPR-TC12 and ESP-GAME, using \textbf{expert labels}.}
\label{tab:comparison_iapr_esp_gt}
\resizebox{0.72\textwidth}{!}{
\begin{tabular}{lcccccccccc}
\toprule
                                                              & \multicolumn{1}{l}{}                 & \multicolumn{4}{c}{\textbf{IAPR-TC12}}                                                                                                      && \multicolumn{4}{c}{\textbf{ESP-GAME}}                                                                                                       \\ 
\cmidrule{3-6} \cmidrule{8-11}                                        
\multicolumn{1}{l}{\textbf{Method}}                                 & \multicolumn{1}{c}{\textbf{Visual Feat}} & \multicolumn{1}{c}{\textbf{MAP}} & \multicolumn{1}{c}{\textbf{Prec@5}} &  \multicolumn{1}{c}{\textbf{Rec@5}} & \multicolumn{1}{c}{\textbf{N+}}           && \multicolumn{1}{c}{\textbf{MAP}} & \multicolumn{1}{c}{\textbf{Prec@5}} & \multicolumn{1}{c}{\textbf{Rec@5}} & \multicolumn{1}{c}{\textbf{N+}}           \\  \cmidrule{1-11}
\multicolumn{11}{l}{\textbf{\emph{State of the art:}}}\\ [3pt] 
\multicolumn{1}{l}{MBRM \cite{feng-2004}}         & \multicolumn{1}{c}{HC}              & - & 24                       & 23                      & \multicolumn{1}{c}{223}          && -  & -                        & -                       & \multicolumn{1}{c}{-}            \\
\multicolumn{1}{l}{JEC-15 \cite{makadia-2008}}   & \multicolumn{1}{c}{HC}              & - & 29                       & 19                      & \multicolumn{1}{c}{211}          && - & -                        & -                       & \multicolumn{1}{c}{-}            \\
\multicolumn{1}{l}{TagProp \cite{guillaumin-2009}} & \multicolumn{1}{c}{HC}              & \textbf{40} & 46                       & 35                      & \multicolumn{1}{c}{266}          && \textbf{28} & 39                       & 27                      & \multicolumn{1}{c}{239}          \\
\multicolumn{1}{l}{GS \cite{zhang-2010}}          & \multicolumn{1}{c}{HC}              & - & 32                       & 29                      & \multicolumn{1}{c}{252}          && - & -                        & -                       & \multicolumn{1}{c}{-}            \\
\multicolumn{1}{l}{RF-opt \cite{fu-2012}}         & \multicolumn{1}{c}{HC}              & - & 44                       & 31                      & \multicolumn{1}{c}{253}          && - & -                        & -                       & \multicolumn{1}{c}{-}            \\
\multicolumn{1}{l}{2PKNN-ML \cite{2pknn-2012}}    & \multicolumn{1}{c}{HC}              & - & \textbf{54}              & \textbf{37}                      & \multicolumn{1}{c}{\textbf{278}}          && - & 53                         & {27}                      & \multicolumn{1}{c}{{252}} \\
\multicolumn{1}{l}{KSVM-VT \cite{svmvt-2013}}     & \multicolumn{1}{c}{HC}              & - & 47                       & 29                      & \multicolumn{1}{c}{268}          && - & \textbf{55}              & 25                      & \multicolumn{1}{c}{\textbf{259}}          \\
\multicolumn{1}{l}{SKL-CRM \cite{moran-2014}}              & \multicolumn{1}{c}{HC}              & - & 47                       & 32                      & \multicolumn{1}{c}{274}          && - & 41                       & 26                      & \multicolumn{1}{c}{248}          \\
\multicolumn{1}{l}{CCA-KNN \cite{murthy-2015}}             & \multicolumn{1}{c}{VGG16}           & - & 41                       & 34                      & \multicolumn{1}{c}{273}          && - & 44                       & \textbf{32}                      & \multicolumn{1}{c}{254}          \\ 

\multicolumn{1}{l}{RLR \cite{izadinia2015deep}}             & \multicolumn{1}{c}{Alexnet}           & - & 46                       & \textbf{41}                      & \multicolumn{1}{c}{277}          && - & - & - & \multicolumn{1}{c}{-}          \\ \midrule

\multicolumn{11}{l}{\textbf{\emph{Baselines:}}}\\ [3pt] 

\multicolumn{1}{l}{NNvot}                       & \multicolumn{1}{c}{VGG16}           & 36 & 39                       & 29                      & \multicolumn{1}{c}{239}          && 28 & 31                       & 28                      & \multicolumn{1}{c}{232}          \\
\multicolumn{1}{l}{TagRel}                      & \multicolumn{1}{c}{VGG16}           & 35 & 34                       & 35                      & \multicolumn{1}{c}{262}          && 30 & 29                       & 31                      & \multicolumn{1}{c}{240}          \\
\multicolumn{1}{l}{TagProp}                     & \multicolumn{1}{c}{VGG16}           & 38 & 40                       & 32                      & \multicolumn{1}{c}{257}          && 32 & 34                       & 32                      & \multicolumn{1}{c}{241}          \\
\multicolumn{1}{l}{2PKNN}                       & \multicolumn{1}{c}{VGG16}           & \textbf{41} & \textbf{41}                       & \textbf{39}             & \multicolumn{1}{c}{\textbf{276}} && \textbf{36} & \textbf{43}                       & \textbf{36}             & \multicolumn{1}{c}{\textbf{257}}          \\
\multicolumn{1}{l}{SVM}                         & \multicolumn{1}{c}{VGG16}           & 34 & 31                       & 29                      & \multicolumn{1}{c}{221}          && 31 & 29                       & 30                      & \multicolumn{1}{c}{224}          \\ \midrule 

\multicolumn{11}{l}{\textbf{\emph{Our Approach:}}}\\ [3pt] 

\multicolumn{1}{l}{KCCA + NNvot}                       & \multicolumn{1}{c}{VGG16}           & 40 & 44                       & 34                      & \multicolumn{1}{c}{250}          && 34 & 38                       & 34                      & \multicolumn{1}{c}{240}          \\
\multicolumn{1}{l}{KCCA + TagRel}                      & \multicolumn{1}{c}{VGG16}           & 40 & 41                       & 37                      & \multicolumn{1}{c}{259}          && 35 & 33                       & 37                      & \multicolumn{1}{c}{249}          \\
\multicolumn{1}{l}{KCCA + TagProp}                     & \multicolumn{1}{c}{VGG16}           & 41 & 44                       & 34                      & \multicolumn{1}{c}{257}          && 37 & 38                       & 36                      & \multicolumn{1}{c}{247}          \\
\multicolumn{1}{l}{KCCA + 2PKNN}                       & \multicolumn{1}{c}{VGG16}           & \textbf{43} & \textbf{49}                       & \textbf{38}             & \multicolumn{1}{c}{\textbf{278}} && \textbf{39} & \textbf{45}                       & \textbf{39}             & \multicolumn{1}{c}{\textbf{260}}          \\
\multicolumn{1}{l}{KCCA + SVM}                         & \multicolumn{1}{c}{VGG16}           & 41 & 44                       & 35                      & \multicolumn{1}{c}{252}          && 37 & 38                       & 37                      & \multicolumn{1}{c}{251}          \\ 
\bottomrule
\end{tabular}
}
\end{table*}

\subsection{Evaluation Protocol}\label{measures}
The performance of automatic image annotation on these datasets has been measured with different metrics. Therefore, for each dataset, we carefully follow previous work protocols.
We employ four popular metrics to assess the performance of our algorithm and compare to existing approaches.

Image annotation is usually addressed by predicting a fixed number of labels, $n$, per image (e.g. $n=3$, $n=5$).
We compute precision (Prec@$n$) and recall (Rec@$n$) by averaging these two metrics over all the labels.
Considering that image ground-truth labels may be less or more than $n$, and we are constrained by this setup to predict $n$ labels, perfect precision and recall can not be obtained.
We also report results using Mean Average Precision (MAP), which takes into account all labels for every image, and evaluates the full ranking.
First, we rank all test images according to the predicted relevance to compute AP for each label, then we report the mean value of AP over all labels.
Finally we report N+ which is often used to denote the number of labels with non-zero recall. N+ is an interesting metric when the set of labels has a moderate to high cardinality, otherwise it tends to saturate easily not providing adequate information on a method.
It has to be noted that each metric evaluates very different properties of each method. Therefore a method hardly dominates over the competition on every metric. Some methods, by design, provide better Recall or Precision than others.

For IAPR-TC12 and ESP-GAME, the standard protocol is to report Prec@5, Rec@5 and N+~\cite{duygulu-2002,makadia-2008}.
For completeness we report MAP on these two datasets although, as can be seen in Table~\ref{tab:comparison_iapr_esp_gt}, few previous work also report this metric.

For MIRFlickr, considering that annotated labels are used to perform image retrieval, the few existing works report only the MAP \cite{verbeek-2010}. We also report Prec@5 and Rec@5. Considering the low cardinality of the tag vocabulary ($18$), N+ is not reported for this dataset.

For NUS-WIDE, performances are usually reported either as MAP or precision and recall. Since NUS-WIDE has a lower average number of labels per image than IAPR-TC12 and ESP-GAME, we report results with $n=3$ labels, as in \cite{gong2013deep,johnson-2015}.

\subsection{Implementation Details and Baselines}
In order to avoid degeneracy with non-invertible Gram matrices and to increase computational efficiency, we approximate the Gram matrices using the Partial Gram-Schmidt Orthogonalization (PGSO) algorithm provided by Hardoon \etal \cite{hardoon-2004}.
In all the experiments we have empirically fixed $\kappa = 0.5$ (see Eq. \ref{eq:kcca_regularization}) since it gave the best performance in early experiments on IAPR-TC12. We use approximate kernel matrices given by the PGSO algorithm, where we consider at most $4,096$ dimensions (i.e. the dimension of the semantic space). Thus the dimensionality of $\psi(I)$ in Eq. \ref{eq:semantic_space} is $4,096$. In this case, the distance between two images is defined as the cosine distance between $\psi$ features. 

Since our approach is based on semantic space built from visual data and the available labels, we consider as baselines the label transfer methods trained on the bare visual features.
The distance between two images $I_q$ and $I_i$ is defined as $d(I_q, I_i) = 1-K^V (I_q, I_i)$, where $K^V$ is the visual kernel described in Eq.~\ref{eq:kernel_visual}, normalized with values in $[0,1]$.

The number of nearest neighbors $K$ and the $C$ of SVM were fixed by performing a 3-fold cross-validation on the training set for each dataset.

\subsection{Experiment 1: Performance with Expert Labels}\label{sec:results_gt}

\begin{table}[t!]
\centering
\caption{Results of our method compared to the state of the art on the dataset MIRFlickr-25K, using \textbf{expert labels}.}
\label{tab:comparison_mirflickr_gt}
\resizebox{0.92\columnwidth}{!}{
\begin{tabular}{lcccc}
\toprule
                                                     & & \multicolumn{3}{c}{\textbf{MIRFlickr-25K}}                                                        \\
\cmidrule{3-5}
\multicolumn{1}{l}{\textbf{Methods}}                        & \multicolumn{1}{c}{\textbf{Visual Feat}} & \textbf{MAP} & \textbf{Prec@5} & \multicolumn{1}{c}{\textbf{Rec@5}}     \\
\midrule
\multicolumn{5}{l}{\textbf{\emph{State of the art:}}}\\ [3pt] 
\multicolumn{1}{l}{TagProp \cite{verbeek-2010}} & \multicolumn{1}{c}{HC}          & 46.5  & - & -                  \\
\multicolumn{1}{l}{SVM \cite{verbeek-2010}}            & \multicolumn{1}{c}{HC}          & 52.3 & - & -   \\
\multicolumn{1}{l}{Autoencoder \cite{srivastava2012dbm}} & \multicolumn{1}{c}{HC}          & 60.0 & - & -                  \\
\multicolumn{1}{l}{DBM \cite{srivastava2012dbm}} & \multicolumn{1}{c}{HC}          & 60.9  & - & -                  \\
\multicolumn{1}{l}{MKL \cite{guillaumin-2010}} & \multicolumn{1}{c}{HC}          & \textbf{62.3}  & - & -                  \\
\midrule
\multicolumn{5}{l}{\textbf{\emph{Baselines:}}}\\ [3pt] 
\multicolumn{1}{l}{NNvot \hide{\cite{makadia-2008}} }                          & \multicolumn{1}{c}{VGG16}       & 69.9   &  44.7 &  69.2                   \\
\multicolumn{1}{l}{TagRel \hide{\cite{xli-2009}}}                         & \multicolumn{1}{c}{VGG16}       & 68.9   & 41.5 & 72.1                        \\
\multicolumn{1}{l}{TagProp \hide{\cite{guillaumin-2009}} }                       & \multicolumn{1}{c}{VGG16}       & 70.8    & \textbf{45.5}  & 70.1                       \\
\multicolumn{1}{l}{2PKNN \hide{\cite{2pknn-2012}} }                         & \multicolumn{1}{c}{VGG16}       & 66.5  & 46.4  &  70.9                         \\
\multicolumn{1}{l}{SVM \hide{\cite{guillaumin-2009}} }                           & \multicolumn{1}{c}{VGG16}       & \textbf{72.7} & 38.8  &  \textbf{72.4}             \\
\midrule
\multicolumn{5}{l}{\textbf{\emph{Our Approach:}}}\\ [3pt] 
\multicolumn{1}{l}{KCCA + NNvot}              & \multicolumn{1}{c}{VGG16}       & 72.9     & 46.1  & 73.1      \\
\multicolumn{1}{l}{KCCA + TagRel}             & \multicolumn{1}{c}{VGG16}       & 70.7    & 45.2  &  72.6                \\
\multicolumn{1}{l}{KCCA + TagProp}            & \multicolumn{1}{c}{VGG16}       & 73.0    & 44.6  & 74.1          \\
\multicolumn{1}{l}{KCCA + 2PKNN}              & \multicolumn{1}{c}{VGG16}       & 67.7   & \textbf{47.3}  & 74.6                   \\
\multicolumn{1}{l}{KCCA + SVM}                & \multicolumn{1}{c}{VGG16}       & \textbf{73.0}   & 38.9  & \textbf{75.0}                     \\
\bottomrule
\end{tabular}
}
\end{table}

\begin{table}[t!]
\centering
\caption{Results on the NUS-WIDE dataset using \textbf{expert labels}.}
\label{tab:comparison_nuswide_gt}
\resizebox{0.95\columnwidth}{!}{
\begin{tabular}{lcccc}
\toprule
                                                                     & & \multicolumn{3}{c}{\textbf{NUS-WIDE}}                                                                                      \\
\cmidrule{3-5}
\multicolumn{1}{l}{\textbf{Methods}}                                        & \multicolumn{1}{c}{\textbf{Visual Feat}}  & \textbf{MAP} & \textbf{Prec@3} & \multicolumn{1}{c}{\textbf{Rec@3}} \\
\midrule
\multicolumn{5}{l}{\textbf{\emph{State of the art:}}}\\ [3pt] 
\multicolumn{1}{l}{CNN + SoftMax \cite{gong2013deep}}     & \multicolumn{1}{c}{RGB}               & -    & 31.7   & \multicolumn{1}{c}{31.2}      \\
\multicolumn{1}{l}{CNN + WARP  \cite{gong2013deep}}        & \multicolumn{1}{c}{RGB}               & -    & 31.7   & \multicolumn{1}{c}{35.6}      \\
\multicolumn{1}{l}{CNN + NNvot \cite{johnson-2015}}     & \multicolumn{1}{c}{BLVC}            & 44.0   & \textbf{44.4} & \multicolumn{1}{c}{30.8}    \\
\multicolumn{1}{l}{CNN + logistic \cite{johnson-2015}}  & \multicolumn{1}{c}{BLVC}            & \textbf{45.8}  & 40.9 & \multicolumn{1}{c}{\textbf{43.1}}   \\
\multicolumn{1}{l}{MIE Ranking \cite{ren-2015}}        & \multicolumn{1}{c}{BLVC}            & -       & 37.9     & \multicolumn{1}{c}{38.9}      \\
\multicolumn{1}{l}{MIE Full Model \cite{ren-2015}}     & \multicolumn{1}{c}{BLVC}            & -      & 37.8   & \multicolumn{1}{c}{40.2}        \\ 
\midrule
\multicolumn{5}{l}{\textbf{\emph{Baselines:}}}\\ [3pt] 
\multicolumn{1}{l}{NNvot \hide{\cite{makadia-2008}}}                                          & \multicolumn{1}{c}{VGG16}           & 49.3       & 39.6  & \multicolumn{1}{c}{44.0}   \\
\multicolumn{1}{l}{TagRel \hide{\cite{xli-2009}} }                                        & \multicolumn{1}{c}{VGG16}           & 49.2     & 32.1 & \multicolumn{1}{c}{50.3} \\
\multicolumn{1}{l}{TagProp \hide{\cite{guillaumin-2009}}  }                                      & \multicolumn{1}{c}{VGG16}           & \textbf{50.9}     & \textbf{41.3} & \multicolumn{1}{c}{44.6}  \\
\multicolumn{1}{l}{2PKNN \hide{\cite{2pknn-2012}} }                                         & \multicolumn{1}{c}{VGG16}           & 48.0    & 39.7 & \multicolumn{1}{c}{52.2}  \\
\multicolumn{1}{l}{SVM \hide{\cite{guillaumin-2009}} }                                           & \multicolumn{1}{c}{VGG16}           & 50.2     & 34.6   & \multicolumn{1}{c}{\textbf{60.6}} \\
\midrule
\multicolumn{5}{l}{\textbf{\emph{Our Approach:}}}\\ [3pt] 
\multicolumn{1}{l}{KCCA + NNvot}                              & \multicolumn{1}{c}{VGG16}           & 51.7     & 40.2   & \multicolumn{1}{c}{50.5}  \\
\multicolumn{1}{l}{KCCA + TagRel}                             & \multicolumn{1}{c}{VGG16}           & 51.4    &  34.4   & \multicolumn{1}{c}{\textbf{57.2}}  \\
\multicolumn{1}{l}{KCCA + TagProp}                            & \multicolumn{1}{c}{VGG16}           & \textbf{52.2}     & 45.2    & \multicolumn{1}{c}{49.2}   \\
\multicolumn{1}{l}{KCCA + 2PKNN}                              & \multicolumn{1}{c}{VGG16}           & 50.7    & \textbf{53.0}      & \multicolumn{1}{c}{47.0}     \\
\multicolumn{1}{l}{KCCA + SVM}                                & \multicolumn{1}{c}{VGG16}           & 51.8    & 43.3    & \multicolumn{1}{c}{48.4}       \\ 
\bottomrule
\end{tabular}
}
\end{table}

As a first experiment we analyze the performance of our method when the semantic space is built from expert labels.
In Tables \ref{tab:comparison_iapr_esp_gt}, \ref{tab:comparison_mirflickr_gt} and \ref{tab:comparison_nuswide_gt} we report the performance of the state of the art, the five methods ran in the visual feature space and in the semantic space, respectively.
Our best result is superior to the state of the art on NUS-WIDE and MIRFlickr-25K while it is comparable to more tailored methods on IAPR-TC12 and ESP-GAME.

Table \ref{tab:comparison_iapr_esp_gt} shows the performance of the state of the art methods, the baselines and our approach on IAPR-TC12 and ESP-GAME. 
We first note that the majority of previous works report results with 15 handcrafted features (HC) \cite{guillaumin-2009} while we use the more recent VGG16 CNN activations, the same as \cite{murthy-2015}.
By exploiting this feature, simple nearest neighbor methods like NNvot and TagRel reach a higher Prec@5 and Rec@5 compared to the similar JEC-15 \cite{makadia-2008} which uses a combination of HC features.
Our baseline TagProp has a slight inferior performance to that reported in \cite{guillaumin-2009}, probably due to the lower number of learnable parameters, having only one single feature versus $15$.
Comparing our approach versus the baselines, we observe that all metrics consistently report higher values when label transfer is applied in the semantic space.
This suggests that classes in the semantic space are easier to separate. We reach our best result on IAPR-TC12 and ESP-GAME with KCCA + 2PKNN, still inferior to 2PKNN-ML \cite{2pknn-2012} that is additionally applying metric learning.

\begin{figure}[!t]
\centering
\includegraphics[width=0.85\columnwidth]{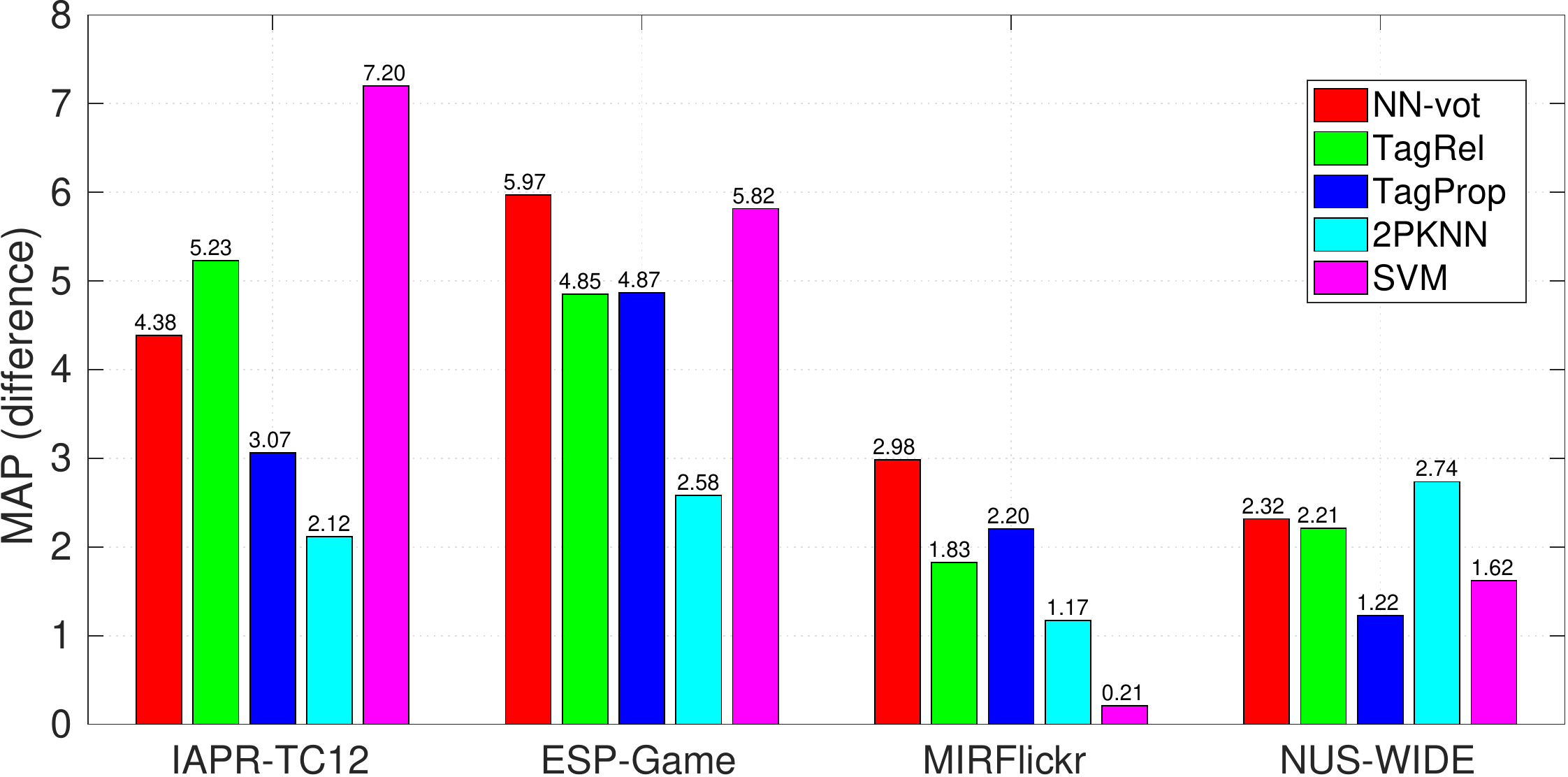}
\caption{MAP difference of the four methods trained with KCCA on ESPGame, IAPR-TC12, MIRFlickr-25k and NUS-WIDE. KCCA is trained using \textbf{expert labels}.}
\label{fig:diffs_performance}
\end{figure}

Table \ref{tab:comparison_mirflickr_gt} shows our results on the MIRFlickr-25k dataset.
Again, we first note that by simply switching from HC features to VGG16, a large boost of MAP is obtained.
Focusing on TagProp and SVM baselines, which are directly comparable with previous work \cite{verbeek-2010}, MAP increases from $52.3$ to $72.7$ and from $46.5$ to $70.8$, respectively. This is consistent with recent literature that suggests CNN activations are way more powerful than handcrafted features.
We also report the experimental results of \cite{srivastava2012dbm}, obtained using autoencoders and multimodal Deep Boltzmann Machines, and \cite{guillaumin-2010} (semi-supervised multimodal kernel learning), which are the previous state-of-the-art results on this dataset.
Applying our KCCA-based framework to the five methods results in a generalized improvement of all metrics, especially on the four nearest neighbor schemes.
The best MAP is obtained by KCCA $+$ SVM that reaches a score of $73.0$, higher than the best baseline. Interestingly, KCCA $+$ NNvot and KCCA $+$ TagProp reach a score of $72.9$, that is higher than the best baseline SVM. We can observe that our semantic space improves both Rec@5 and Prec@5, specifically an average increase of 3.1 for Rec@5 and of 2.1 of Prec@5 can be measured for all 5 baseline methods.

We report in Table \ref{tab:comparison_nuswide_gt} the results of the comparison on the large-scale NUS-WIDE dataset.
Previous works used BLVC (Caffe reference model) features (e.g. \cite{johnson-2015}) while we use VGG16, but this does not provide significant differences in performance.
Moreover Gong~\etal \cite{gong2013deep} attempted to train the network from scratch, obtaining an inferior performance with respect to pre-trained features on ImageNet \cite{johnson-2015,gong2013deep}.
A higher score of Rec@3 is observed in all our experiments with respect to the state of the art. This suggests that our approach is able to work with unbalanced distribution of labels, and improves recall of rare labels.
KCCA $+$ TagProp is the overall best method on this dataset, even superior to SVM that is commonly recognized as better than kNN-based methods for classification. 

In summary, our framework is always able to improve performance in all datasets with every metric.
This is an important result since each particular metric captures different properties.
On smaller datasets, such as IAPR-TC12 and ESP-GAME, metric learning based approaches~\cite{2pknn-2012,guillaumin-2009} take more advantage from using 15 different but weaker features then a single, stronger one, as we do. Although on larger and more challenging datasets, such as MIRFlickr and NUS-WIDE, this effect is largely moderated.
Finally, Figure \ref{fig:diffs_performance} shows the difference of MAP between the semantic space and their baseline, for all the five methods.
We highlight that the improvement is generally higher on IAPR-TC12 and ESP-GAME, where fewer training examples are available. In particular, SVM has the largest gain followed by the simpler NNvot and TagRel.
This might be because these methods suffer on rare concepts due to sample insufficiency.


\subsection{Experiment 2: Performance with User Tags}\label{sec:results_tags}

\begin{table*}[!ht]
\centering
\caption{Results on the MIRFlickr-25k and NUS-WIDE datasets using \textbf{user tags}.}
\label{tab:comparison_user_tags}
\resizebox{0.72\textwidth}{!}{
\begin{tabular}{lcccccccc}
\toprule
                                                      & & \multicolumn{3}{c}{\textbf{MIRFlickr-25k}} && \multicolumn{3}{c}{\textbf{NUS-WIDE}}                                            \\
\cmidrule{3-5} \cmidrule{7-9}
\multicolumn{1}{l}{\textbf{Methods}}                         & \multicolumn{1}{c}{\textbf{Visual Feat}} & \textbf{MAP} & \textbf{Prec@5} & \textbf{Rec@5} && \textbf{MAP} & \textbf{Prec@5} & \textbf{Rec@5}   \\
\midrule
\multicolumn{9}{l}{\textbf{\emph{State of the art:}}}\\ [3pt] 
\multicolumn{1}{l}{SVM v \cite{verbeek-2010}}   & \multicolumn{1}{c}{HC}          & 35.4  & - & -  && -  & - & - \\
\multicolumn{1}{l}{SVM v+t \cite{verbeek-2010}} & \multicolumn{1}{c}{HC}          & 37.9  & - & -  && -  & - & -   \\
\multicolumn{1}{l}{TagProp \cite{verbeek-2010}} & \multicolumn{1}{c}{HC}          & 38.4  & - & -  && -  & - & -   \\
\multicolumn{1}{l}{FisherBoxes \cite{uricchio-2015}} & \multicolumn{1}{c}{VGG128}      & \textbf{54.8} & - & - && \textbf{39.7} & - & - \\
\midrule
\multicolumn{5}{l}{\textbf{\emph{Baselines:}}}\\ [3pt] 
\multicolumn{1}{l}{NNVot \hide{\cite{makadia-2008}}}              & \multicolumn{1}{c}{VGG16} & \textbf{59.3} & 34.2 & 67.1 &     & \textbf{43.1}           & 30.1   & \multicolumn{1}{c}{46.3}    \\
\multicolumn{1}{l}{TagRel \hide{\cite{xli-2009}}  }               & \multicolumn{1}{c}{VGG16} & 59.2  & 34.8  &  \textbf{68.0} & & 42.5                     & 27.9 & \multicolumn{1}{c}{49.7}    \\
\multicolumn{1}{l}{TagProp \hide{\cite{guillaumin-2009}}     }    & \multicolumn{1}{c}{VGG16} & 58.1 & 33.5 & 66.0  &        & 42.8          & 28.4 & \multicolumn{1}{c}{\textbf{50.2}}   \\
\multicolumn{1}{l}{2PKNN \hide{\cite{2pknn-2012}}  }              & \multicolumn{1}{c}{VGG16} & 51.4 & 35.9  & 67.1  &   & 41.2                  & \textbf{37.5}    & \multicolumn{1}{c}{43.7}    \\
\multicolumn{1}{l}{SVM \hide{\cite{guillaumin-2009}}  }           & \multicolumn{1}{c}{VGG16} & 43.8 & \textbf{40.0}  & 50.8  &        & 35.5                  & 30.4 & \multicolumn{1}{c}{45.2}    \\
\midrule
\multicolumn{5}{l}{\textbf{\emph{Our Approach:}}}\\ [3pt] 
\multicolumn{1}{l}{KCCA + NNvot}               & \multicolumn{1}{c}{VGG16}       & \textbf{60.6}  & 35.4  & \textbf{68.8}  &   & \textbf{43.7}         & 36.3       & \multicolumn{1}{c}{48.0}     \\
\multicolumn{1}{l}{KCCA + TagRel}              & \multicolumn{1}{c}{VGG16}       & 59.8          & 37.2  & 68.5  &   & 43.5                   & 29.0          & \multicolumn{1}{c}{\textbf{55.1}}\\
\multicolumn{1}{l}{KCCA + TagProp}             & \multicolumn{1}{c}{VGG16}       & 59.7          & 33.6  & 67.4  &   & 42.9                   & 29.3          & \multicolumn{1}{c}{51.3}         \\
\multicolumn{1}{l}{KCCA + 2PKNN}               & \multicolumn{1}{c}{VGG16}       & 56.8          & \textbf{42.9}  & 65.4 &   & 42.0                   & \textbf{56.9} & \multicolumn{1}{c}{34.0}         \\
\multicolumn{1}{l}{KCCA + SVM}                 & \multicolumn{1}{c}{VGG16}       & 47.1          & 37.5  & 56.5  &   & 41.6                   & 37.9          & \multicolumn{1}{c}{47.6}         \\
\bottomrule
\end{tabular}
}
\end{table*}

We now turn our attention to the more difficult setting of noisy user tags. Instead of using expert labels, we rely on user tags as training labels and repeat the same experiments of Section \ref{sec:results_gt}. Only MIRFlickr-25k and NUS-WIDE provide user tags, therefore we report results on these datasets.

Table \ref{tab:comparison_user_tags} shows the performance of the state of the art, the baselines and our approach on MIRFlickr-25k and NUS-WIDE.
As previously noted, changing the features from HC to VGG16 has a strong positive impact.
Comparing the methods ran in the semantic space to the baselines ran on the bare visual feature, we observe that every metric is generally improved.
FisherBoxes \cite{uricchio-2015} uses improved features with the same TagProp algorithm, as our baseline.
Since our TagProp MAP is higher than FisherBoxes, this suggests that VGG16 features alone are more powerful than the combinations of VGG128 boxes.
SVM is inferior to nearest neighbor techniques in terms of MAP while having comparable precision and recall. 
Consistently to expert labels results, 2PKNN performs poorly on NUS-WIDE. In the first phase few images per label are selected, thus reducing its power to address the high visual variability of images with frequent labels. 
We also note that all scores are lower than those reported with expert labels in Table \ref{tab:comparison_mirflickr_gt} and Table \ref{tab:comparison_nuswide_gt}. In particular SVM MAP is the most hampered. This is expected given the noise in user tags, and was also noted in previous work \cite{verbeek-2010}.

\begin{figure}
\centering     
\includegraphics[width=0.85\columnwidth]{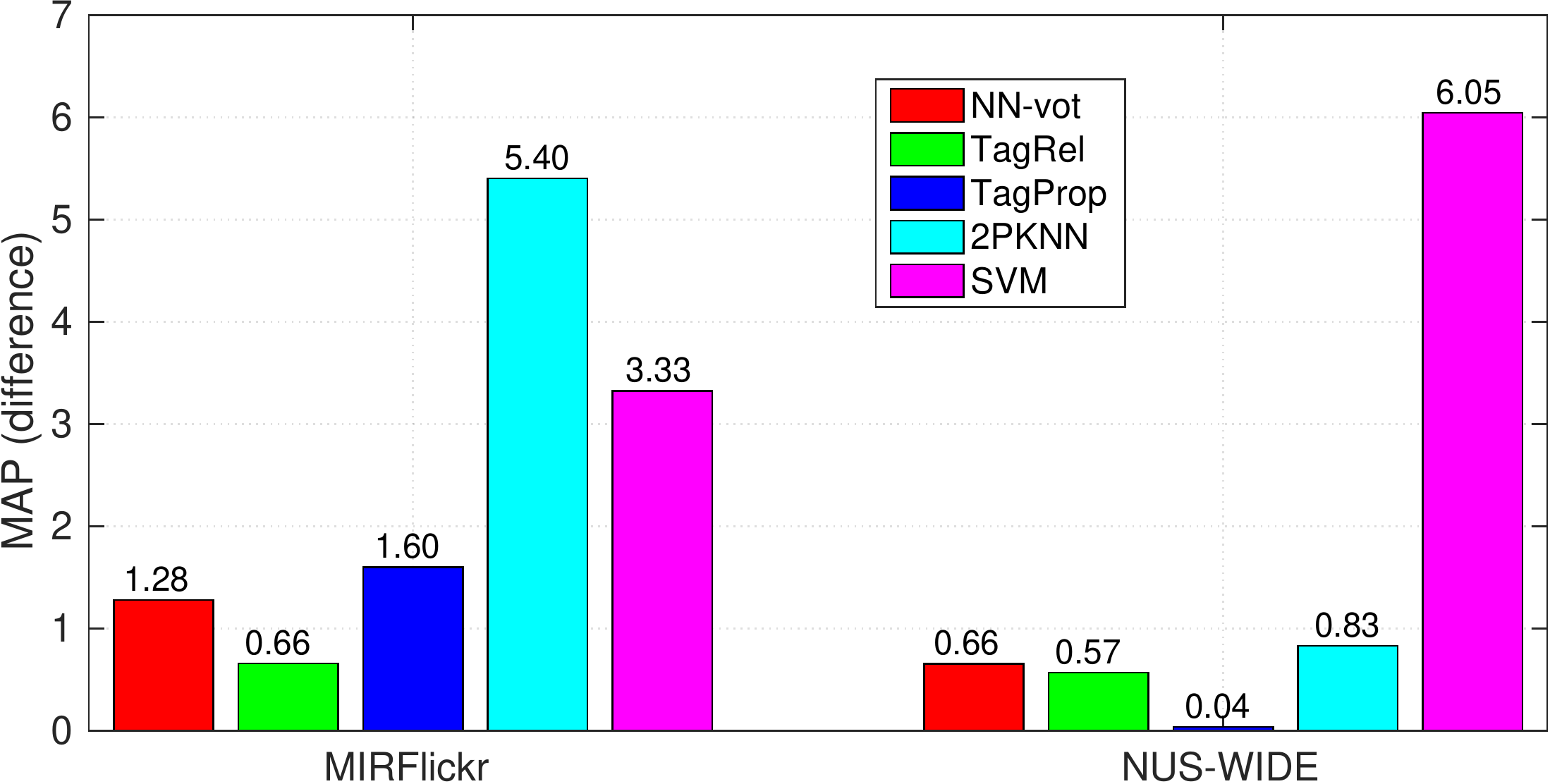}
\caption{MAP relative difference of the four methods trained with KCCA on ESP-Game, IAPR-TC12, MIRFlickr-25k and NUS-WIDE. KCCA is trained with \textbf{user tags}.}
\label{fig:diffs_performance_tags}
\end{figure}

In Figure \ref{fig:diffs_performance_tags} we report the relative MAP difference of the five methods with our technique and the baselines. 
We observe that largest gains are obtained with 2PKNN and SVM. We believe this is due to the fact that 2PKNN and SVM have numerous learning parameters that are likely to generate complex boundaries with label noise.
In contrast, the other three schemes have few or no parameters at all. 
This suggests that features in the semantic space have also some robustness to tag noise.

\begin{table}[!ht]
\centering
\caption{Ablation study on the denoising method. Results are in terms of MAP.}
\label{tab:denoising_mirflickr}
\resizebox{0.95\columnwidth}{!}{
\begin{tabular}{lccc}
\toprule
    & \multicolumn{3}{c}{\textbf{MIRFlickr-25k}}            \\
\cmidrule{2-4}
\multicolumn{1}{l}{\textbf{Methods}}                         & \multicolumn{1}{c}{\textbf{Baseline}} & \textbf{KCCA - NoPreProp}         & \textbf{KCCA} \\
\midrule
\multicolumn{1}{l}{NNvot \cite{makadia-2008} }                    & 59.3             & 56.2     &     \textbf{60.6}     \\
\multicolumn{1}{l}{TagRel \cite{xli-2009} }                       & 59.2             & 54.5     &     \textbf{59.8}     \\
\multicolumn{1}{l}{TagProp \cite{guillaumin-2009} }               & 58.1             & 54.9     &     \textbf{59.7}     \\
\multicolumn{1}{l}{2PKNN \cite{2pknn-2012}  }                     & 51.4             & 42.9     &     \textbf{56.8}     \\
\multicolumn{1}{l}{SVM \cite{guillaumin-2009} }                   & 43.8             & 41.3     &     \textbf{47.1}     \\
\bottomrule
\end{tabular}
}
\end{table}

We believe that such robustness is partially due to the denoising algorithm. To confirm this, we perform an ablation study on MIRFlickr-25k with the same settings as before, except that we omit the pre-propagation step. We report in Table \ref{tab:denoising_mirflickr} the MAP of three different cases: (i) the baseline methods (Baseline); (ii) our approach without the pre-propagation step (KCCA - NoPreProp); (iii) our full approach (KCCA). We observe that avoiding the denoising step leads to an inferior MAP, even less than the baseline case. This confirms that, in presence of excessive sparsity like that in MIRFlickr-25k, KCCA alone is unable to improve the visual features.


\subsection{Experiment 3: Performance with different Textual Features}\label{sec:results_txtkernels}
In this section, we compare the performance of the three proposed textual kernels, defined in Section~\ref{sec:expertlabels}, on expert labels: a bag-of-words linear kernel (\emph{Linear}), a semantic ontology-based kernel (\emph{Ontology}) and a continuous word vector kernel (\emph{Word2Vec}).
Here we perform an experiment with the same settings as experiment 1 (Section~\ref{sec:results_gt}), but the Linear kernel is swapped with the Ontology or Word2Vec kernels. For the Ontology kernel we use WordNet as the underlying ontology while for Word2Vec we employ the pre-trained word vectors on news article.
In Table \ref{tab:comparison_txt_kernels}, we report results on the two largest datasets MIRFlickr-25k and NUS-WIDE, but similar results were obtained on ESP-Game and IAPR-TC12.

\begin{table*}[!ht]
\centering
\caption{Results of our method with the Linear, Ontology and Word2Vec textual kernels on MIRFlickr-25k and NUS-WIDE, using \textbf{expert labels}.}
\label{tab:comparison_txt_kernels}
\resizebox{0.72\textwidth}{!}{
\begin{tabular}{lcccccccc}
\toprule
                                                              & \multicolumn{1}{l}{}                 & \multicolumn{3}{c}{\textbf{MIRFlickr-25k}}                                                                                                      && \multicolumn{3}{c}{\textbf{NUS-WIDE}}                                                                                                       \\ 
\cmidrule{3-5} \cmidrule{7-9}                                        
\multicolumn{1}{l}{\textbf{Method}}                                 & \multicolumn{1}{c}{\textbf{Textual Kernel}} & \multicolumn{1}{c}{\textbf{MAP}} & \multicolumn{1}{c}{\textbf{Prec@5}} &  \multicolumn{1}{c}{\textbf{Rec@5}}          && \multicolumn{1}{c}{\textbf{MAP}} & \multicolumn{1}{c}{\textbf{Prec@5}} & \multicolumn{1}{c}{\textbf{Rec@5}}            \\  \cmidrule{1-9}

\multicolumn{9}{l}{\textbf{\emph{Baselines:}}}\\ [3pt] 

\multicolumn{1}{l}{NNvot}                       & \multicolumn{1}{c}{-} & 69.9 & 44.7 & 69.2 && 49.3  & 39.6  & \multicolumn{1}{c}{44.0}   \\
\multicolumn{1}{l}{TagRel}                      & \multicolumn{1}{c}{-} & 68.9 & 41.5 & 72.1 && 49.2  & 32.1 & \multicolumn{1}{c}{50.3} \\
\multicolumn{1}{l}{TagProp}                     & \multicolumn{1}{c}{-} & 70.8 & 45.5 & 70.1 && \textbf{50.9} & \textbf{41.3} & \multicolumn{1}{c}{44.6}  \\
\multicolumn{1}{l}{2PKNN}                       & \multicolumn{1}{c}{-} & 66.5 & \textbf{46.4} & 70.9 && 48.0  & 39.7 & \multicolumn{1}{c}{52.2}  \\
\multicolumn{1}{l}{SVM}                         & \multicolumn{1}{c}{-} & \textbf{72.7} & 38.8 & \textbf{72.4} && 50.2  & 34.6   & \multicolumn{1}{c}{\textbf{60.6}} \\ \midrule

\multicolumn{9}{l}{\textbf{\emph{Our Approach:}}}\\ [3pt] 

\multicolumn{1}{l}{KCCA + NNvot}                       & \multicolumn{1}{c}{Linear}           & \textbf{72.9} & 46.1  & 73.1 && \textbf{51.7}     & 40.2   & \multicolumn{1}{c}{\textbf{50.5}} \\
\multicolumn{1}{l}{KCCA + NNvot}                       & \multicolumn{1}{c}{Ontology}         & 72.5 & 46.6 & 72.3 && 51.2 & \textbf{46.7}	& 46.3 \\
\multicolumn{1}{l}{KCCA + NNvot}                       & \multicolumn{1}{c}{Word2Vec}         & 72.3 & \textbf{46.9} & \textbf{73.4} && 50.6 & 40.8	& 50.1 \\ \cmidrule{1-9}
\multicolumn{1}{l}{KCCA + TagRel}                      & \multicolumn{1}{c}{Linear}           & 70.7    & 45.2  &  72.6 && \textbf{51.4}    &  34.4   & \multicolumn{1}{c}{\textbf{57.2}} \\
\multicolumn{1}{l}{KCCA + TagRel}                      & \multicolumn{1}{c}{Ontology}         & 70.6 & \textbf{47.4} & 73.9          && 49.5 & \textbf{35.9}	& 54.3 \\
\multicolumn{1}{l}{KCCA + TagRel}                      & \multicolumn{1}{c}{Word2Vec}         & \textbf{70.9} & 47.2 & \textbf{74.2} && 49.8 & 34.9	& 57.0 \\ \cmidrule{1-9}
\multicolumn{1}{l}{KCCA + TagProp}                     & \multicolumn{1}{c}{Linear}           & \textbf{73.0}    & 44.6  & \textbf{74.1} && \textbf{52.2}     & \textbf{45.2}    & \multicolumn{1}{c}{49.2}    \\
\multicolumn{1}{l}{KCCA + TagProp}                     & \multicolumn{1}{c}{Ontology}         & 72.7 & 44.6 & 73.7 && 51.7 & \textbf{45.2}	& 48.1 \\
\multicolumn{1}{l}{KCCA + TagProp}                     & \multicolumn{1}{c}{Word2Vec}         & 72.9 & \textbf{45.3} & 73.8 && 51.6 & 40.9	& \textbf{50.6} \\ \cmidrule{1-9}
\multicolumn{1}{l}{KCCA + 2PKNN}                       & \multicolumn{1}{c}{Linear}           & \textbf{67.7}   & \textbf{47.3}  & 74.6 && \textbf{50.7}    & \textbf{53.0}      & \multicolumn{1}{c}{47.0} \\
\multicolumn{1}{l}{KCCA + 2PKNN}                       & \multicolumn{1}{c}{Ontology}         & 65.7 & 44.1 & \textbf{76.1} && 49.2 & 46.3	& 51.1 \\
\multicolumn{1}{l}{KCCA + 2PKNN}                       & \multicolumn{1}{c}{Word2Vec}         & 66.2 & 44.2 & 75.7 && 48.9 & 47.3	& \textbf{51.4} \\ \cmidrule{1-9}
\multicolumn{1}{l}{KCCA + SVM}                         & \multicolumn{1}{c}{Linear}           & \textbf{73.0}   & 38.9  & \textbf{75.0} && \textbf{51.8}    & 43.3    & \multicolumn{1}{c}{\textbf{48.4}} \\
\multicolumn{1}{l}{KCCA + SVM}                         & \multicolumn{1}{c}{Ontology}         & 71.4 & 39.3 & 73.0 && 51.4 & \textbf{44.7}	& 46.7 \\ 
\multicolumn{1}{l}{KCCA + SVM}                         & \multicolumn{1}{c}{Word2Vec}         & 71.8 & \textbf{39.5} & 74.1 && 50.2 & 42.7	& 47.7 \\ 
\bottomrule
\end{tabular}
}
\end{table*}

We observe that all methods have better performance than the baseline when using our approach, regardless of the textual kernel. 
Some combinations of kernels and methods favor one metric over the others, although the Linear Kernel has almost always the best MAP. Nevertheless, these slight differences in performance do not suggest a superiority of a kernel over the others. We believe that further studies on how to integrate label relations in KCCA are required, leaving the problem of choosing a better textual kernel for KCCA open.


\subsection{Experiment 4: Varying the Size of Neighborhood}

\iftoggle{single}{
\renewcommand{\figwidth}{0.45}
}{
\renewcommand{\figwidth}{0.49}
}

\begin{figure}
\centering     
\includegraphics[width=\figwidth\columnwidth]{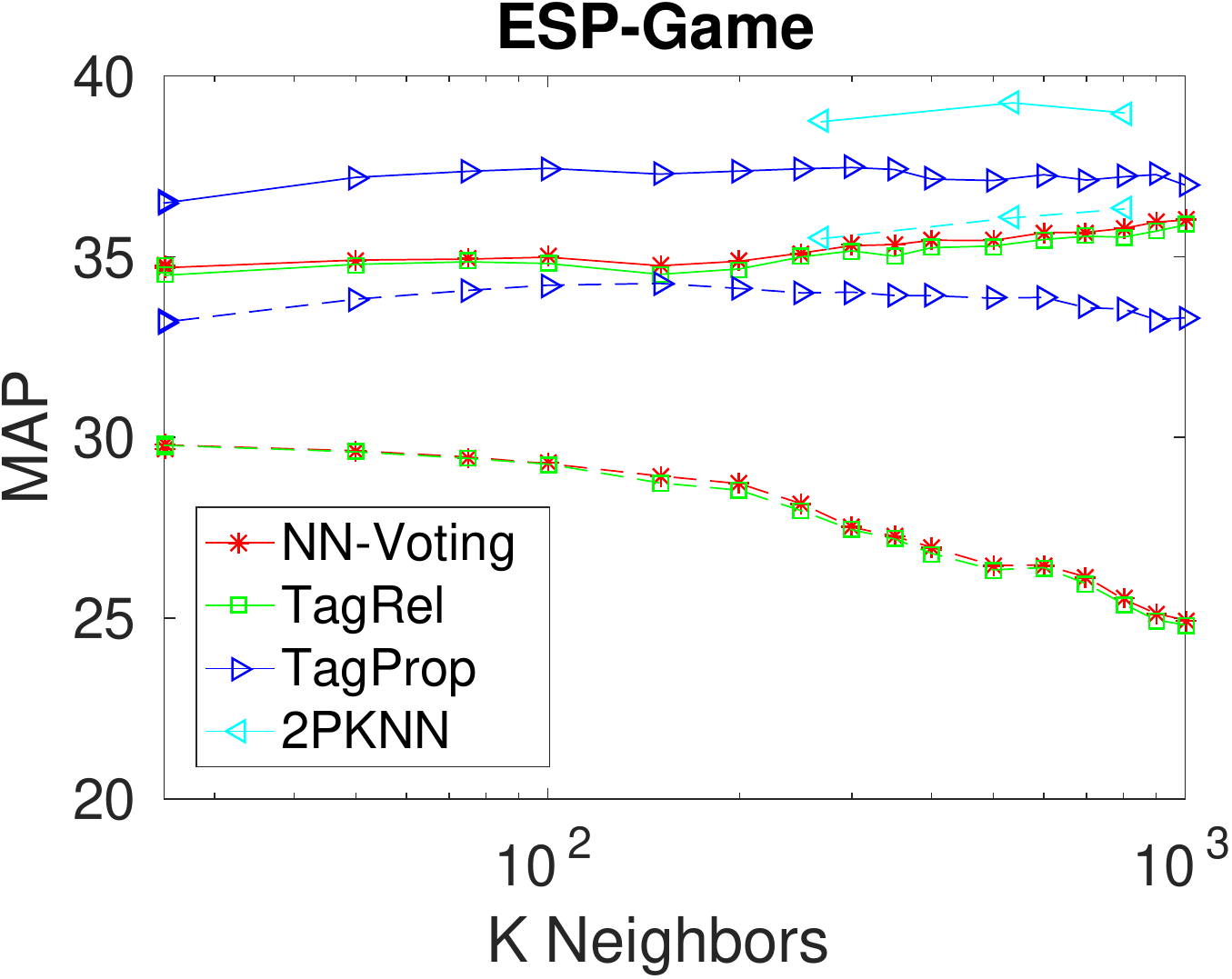}
\includegraphics[width=\figwidth\columnwidth]{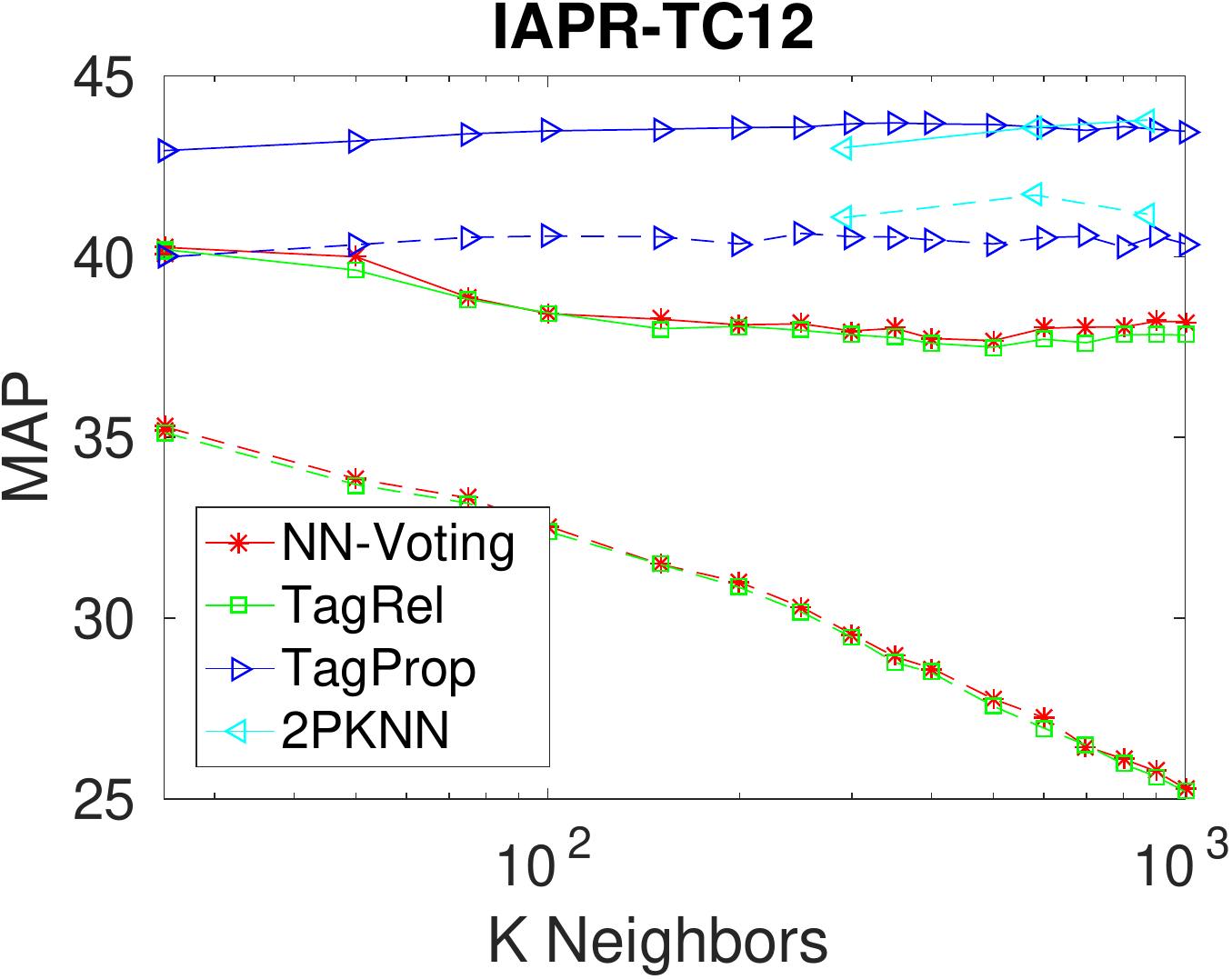}\\
\includegraphics[width=\figwidth\columnwidth]{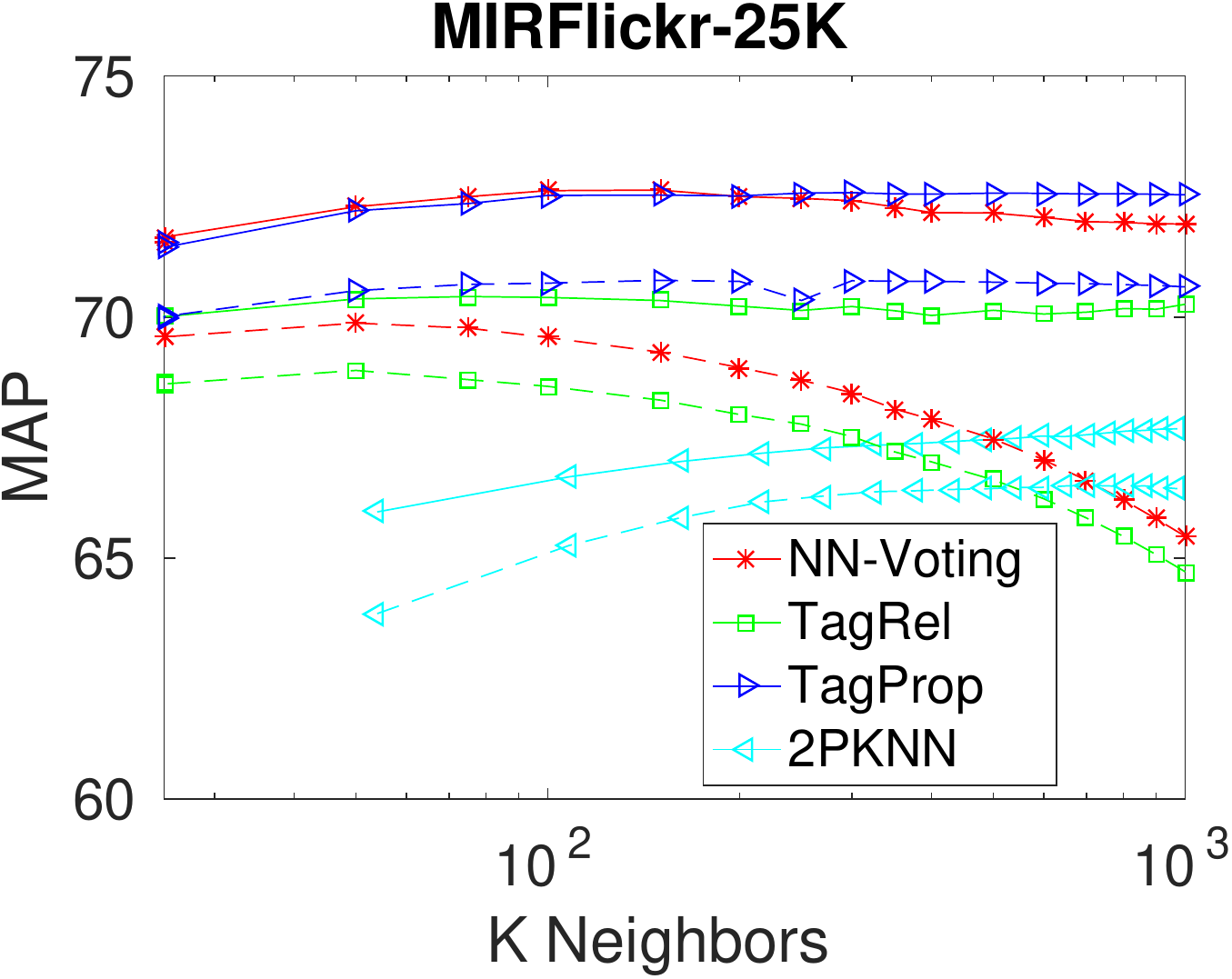}
\includegraphics[width=\figwidth\columnwidth]{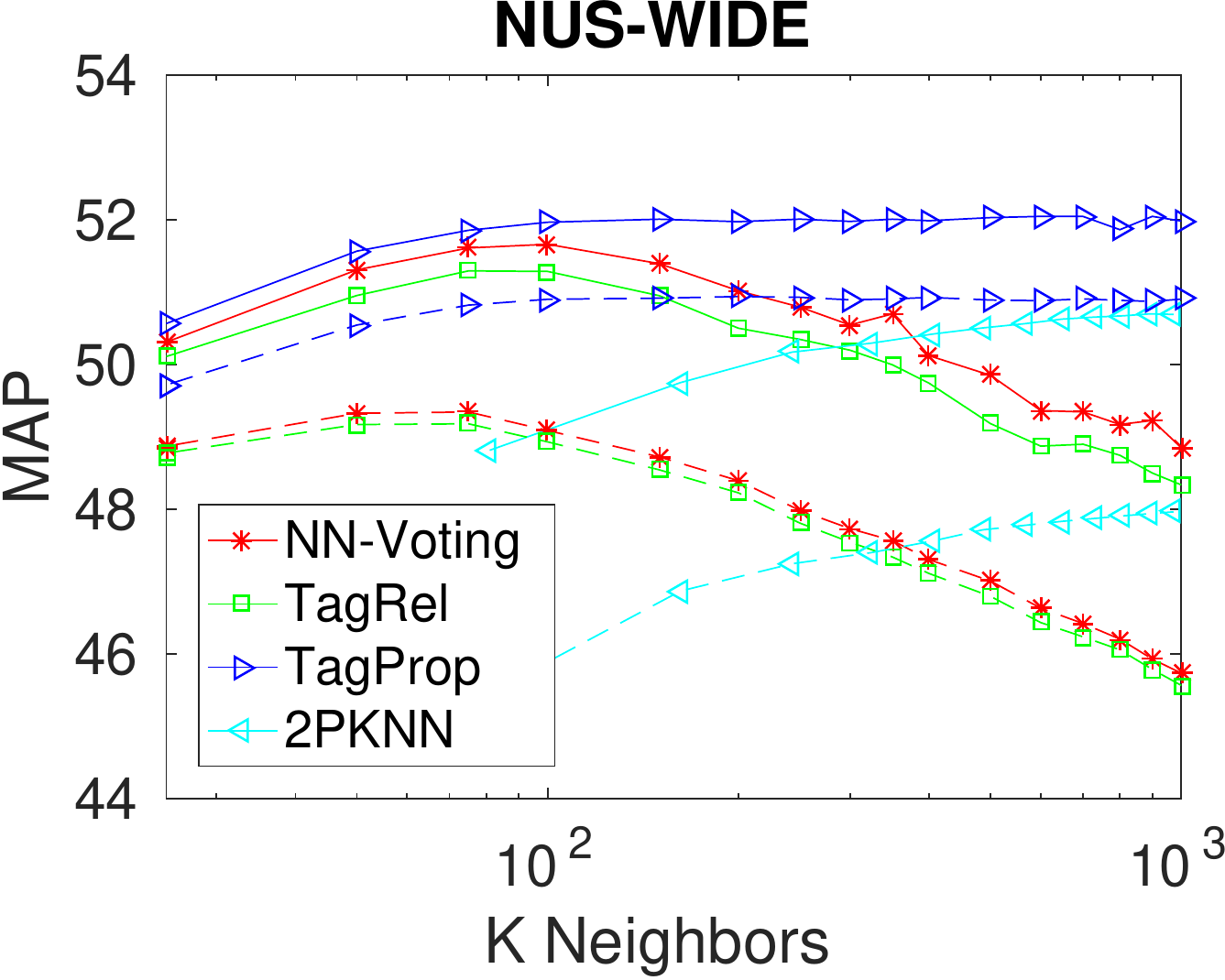}
\caption{MAP of NN-voting, TagRel and TagProp trained with KCCA on ESPGame, IAPR-TC12, MIRFlickr-25k and NUS-WIDE varying the number of nearest neighbors. KCCA is trained with \textbf{expert labels}. Dashed lines represent baseline methods.}
\label{fig:MAP_nearest_neighbor_gt}
\end{figure}

\begin{figure}
\centering     
\includegraphics[width=\figwidth\columnwidth]{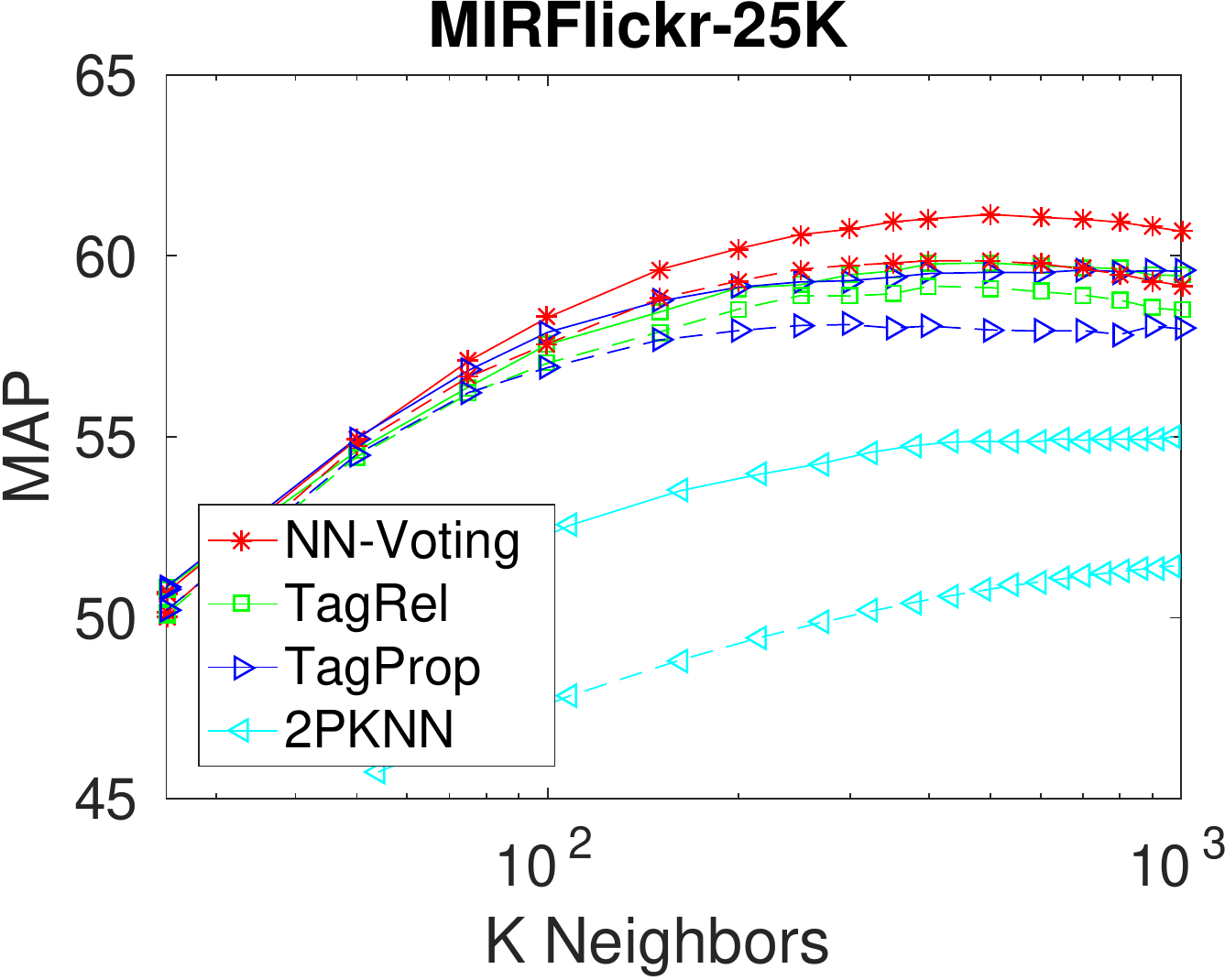}
\includegraphics[width=\figwidth\columnwidth]{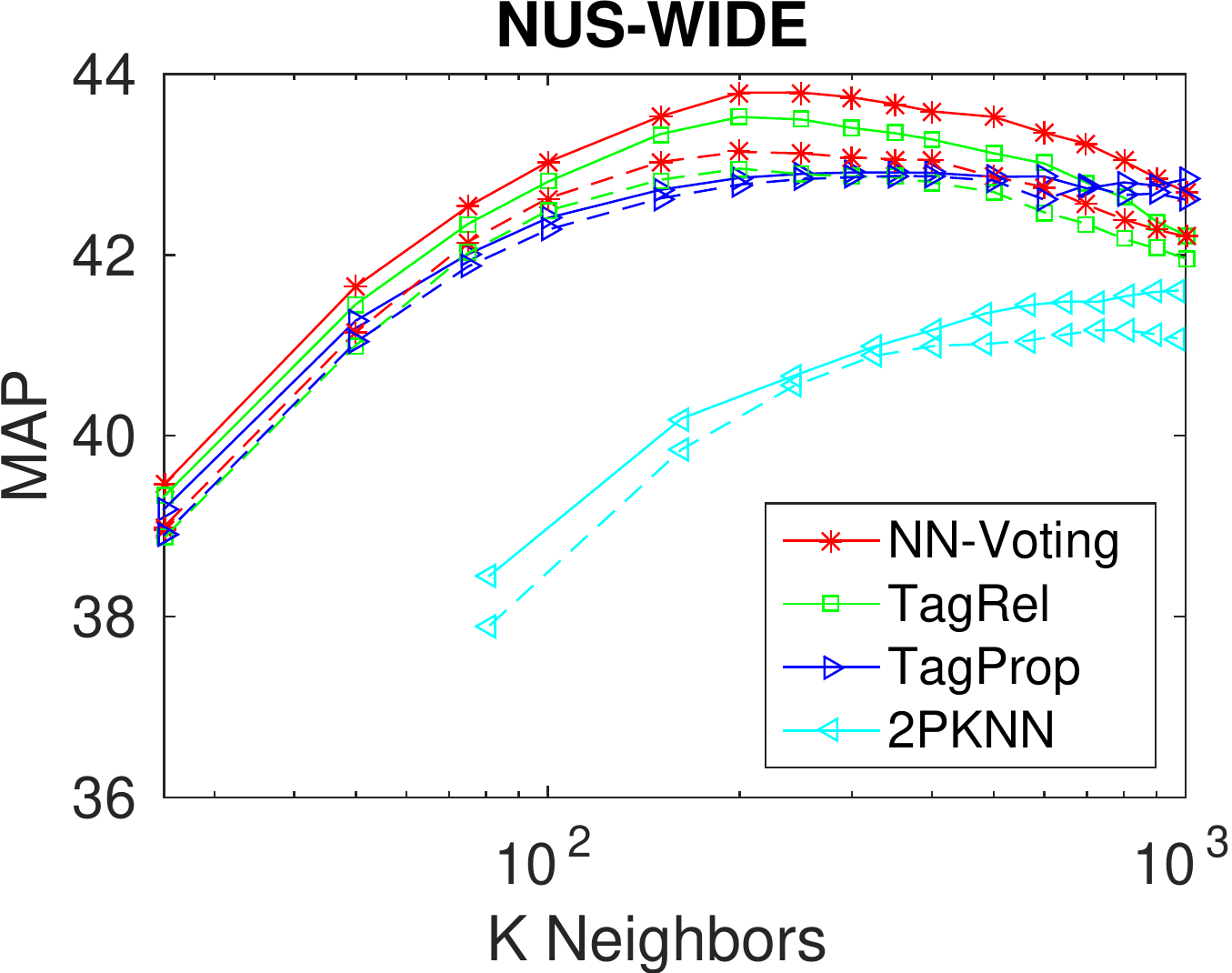}
\caption{MAP evaluation for NN-voting, TagRel and TagProp trained with KCCA on MIRFlickr-25k and NUS-WIDE varying the number of nearest neighbors. KCCA is trained with \textbf{user tags}. Dashed lines represent baseline methods.}
\label{fig:MAP_nearest_neighbor_tags}
\end{figure}

Nearest neighbor methods proved to be well performing on all settings we considered. Although they are simple and do not require much training, they still depend on choosing the right number $K$ of nearest neighbors. 
Thus, we conduct an evaluation of how $K$ affect the performance for both our approach and the baselines. 
Since SVM does not use neighbors, we only perform this evaluation on NNvot, TagRel, TagProp and 2PKNN.

We report in Figures \ref{fig:MAP_nearest_neighbor_gt} and \ref{fig:MAP_nearest_neighbor_tags} the MAP scores when using the expert labels and the user tags, respectively.
As can be seen from both figures, the KCCA variant of the nearest neighbor methods (solid lines) have systematically better MAP than baselines, for any number of neighbors used.
As expected, MAP scores are lower when using user tags (Figure \ref{fig:MAP_nearest_neighbor_tags}). Nevertheless, a gain is observed for each method with any number of neighbors selected. This again confirms that features in the semantic space are better re-arranged, since images with similar semantics are closer in this space.

\begin{figure*}
\centering
\resizebox{0.85\textwidth}{!}{
\begin{tabular}{b{1cm}ccccccccc}
\toprule
                               &                                                                                                    & \multicolumn{2}{c}{\textbf{NNVot}}                                                                                                                                              && \multicolumn{2}{c}{\textbf{TagRel}}                                                                                                                                                    && \multicolumn{2}{c}{\textbf{TagProp}}                                                                                                                                       \\
\cmidrule{3-4} \cmidrule{6-7} \cmidrule{9-10}
\multicolumn{1}{c}{\textbf{Image}}    & \textbf{Exp Labels}                                                                                       & \textbf{Baseline}                                                                          & \textbf{Our}                                                                          && \textbf{Baseline}                                                                             & \textbf{Our}                                                                              && \textbf{Baseline}                                                                        & \textbf{Our}                                                                       \\ \midrule
\multicolumn{1}{c}{\begin{tabular}[c]{@{}c@{}}\includegraphics[width=0.15\textwidth]{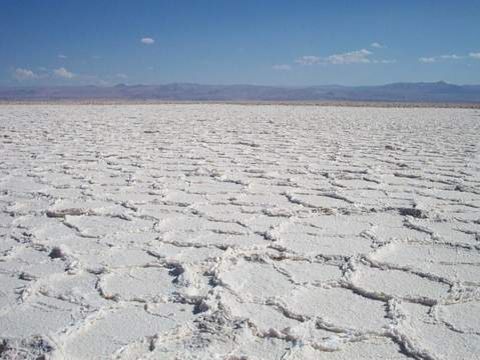}\end{tabular}
} & \begin{tabular}[c]{@{}c@{}}desert\\ mountain\\ range\\ salt\\ sky\end{tabular}                     & \begin{tabular}[c]{@{}c@{}}beach\\ cloud\\ \textbf{mountain}\\ sea\\ \textbf{sky}\end{tabular}      & \begin{tabular}[c]{@{}c@{}}cloud\\ \textbf{desert}\\ landscape\\ \textbf{mountain}\\ \textbf{sky}\end{tabular} && \begin{tabular}[c]{@{}c@{}}beach\\ cloud\\ sea\\ \textbf{sky}\\ wave\end{tabular}             & \begin{tabular}[c]{@{}c@{}}\textbf{desert}\\ lake\\ landscape\\ \textbf{mountain}\\ \textbf{salt}\end{tabular}     && \begin{tabular}[c]{@{}c@{}}beach\\ cloud\\ \textbf{mountain}\\ sea\\ \textbf{sky}\end{tabular}    & \begin{tabular}[c]{@{}c@{}}\textbf{desert}\\ man\\ \textbf{mountain}\\ \textbf{salt}\\ \textbf{sky}\end{tabular}     \\ \midrule
\multicolumn{1}{c}{\begin{tabular}[c]{@{}c@{}}\includegraphics[width=0.16\textwidth]{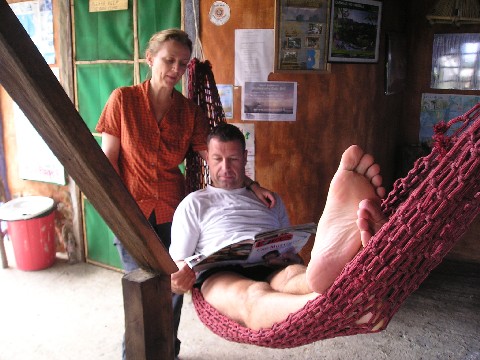}\end{tabular}} & \begin{tabular}[c]{@{}c@{}}hammock\\ man\\ woman\end{tabular}                                      & \begin{tabular}[c]{@{}c@{}}\textbf{man}\\ room\\ table\\ wall\\ \textbf{woman}\end{tabular}         & \begin{tabular}[c]{@{}c@{}}front\\ house\\ \textbf{man}\\ wall\\ \textbf{woman}\end{tabular}          && \begin{tabular}[c]{@{}c@{}}\textbf{man}\\ room\\ table\\ wall\\ \textbf{woman}\end{tabular}            & \begin{tabular}[c]{@{}c@{}}front\\ \textbf{hammock}\\ \textbf{man}\\ wall\\ \textbf{woman}\end{tabular}            && \begin{tabular}[c]{@{}c@{}}bottle\\ \textbf{man}\\ people\\ table\\ \textbf{woman}\end{tabular}   & \begin{tabular}[c]{@{}c@{}}front\\ \textbf{hammock}\\ \textbf{man}\\ wall\\ \textbf{woman}\end{tabular}     \\ \midrule
\multicolumn{1}{c}{\begin{tabular}[c]{@{}c@{}}\includegraphics[width=0.15\textwidth]{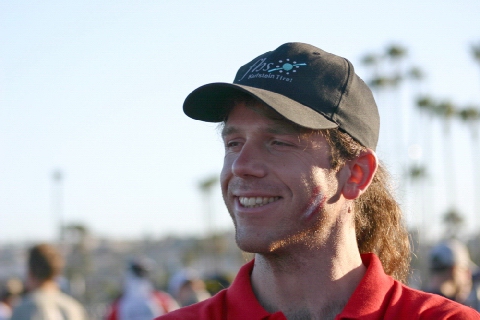}\end{tabular}} & \begin{tabular}[c]{@{}c@{}}cap \\ flag \\ hair \\ man \\ polo \\ portrait \\ shirt\end{tabular}                     & \begin{tabular}[c]{@{}c@{}}boy \\ girl \\ hat \\ \textbf{man} \\ sky\end{tabular}       & \begin{tabular}[c]{@{}c@{}}\textbf{cap} \\ front \\ \textbf{man} \\ sky \\ woman \end{tabular}         && \begin{tabular}[c]{@{}c@{}}boy \\ \textbf{cap} \\ girl \\ \textbf{hair} \\ hat\end{tabular}          & \begin{tabular}[c]{@{}c@{}}boy \\ \textbf{cap} \\ hat \\ \textbf{man} \\ \textbf{shirt}\end{tabular}            && \begin{tabular}[c]{@{}c@{}}boy \\ \textbf{cap} \\ child \\ sky \\ sweater\end{tabular}    & \begin{tabular}[c]{@{}c@{}}\textbf{cap} \\ \textbf{man} \\ \textbf{polo} \\ \textbf{portrait} \\ \textbf{shirt}\end{tabular}    \\ \midrule
\multicolumn{1}{c}{\begin{tabular}[c]{@{}c@{}}\includegraphics[width=0.12\textwidth]{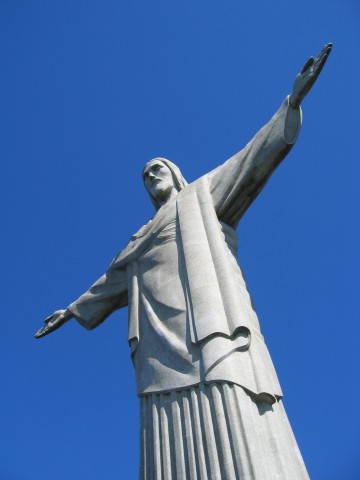}\end{tabular}}  & \begin{tabular}[c]{@{}c@{}}man \\ sky \\ statue \\ view\end{tabular} & \begin{tabular}[c]{@{}c@{}}building \\ front \\ people \\ \textbf{sky} \\ tower \end{tabular} & \begin{tabular}[c]{@{}c@{}}front \\ \textbf{man} \\ \textbf{sky} \\ \textbf{statue} \\ tree\end{tabular} && \begin{tabular}[c]{@{}c@{}}building \\ column \\ \textbf{sky} \\ \textbf{statue} \\ tower\end{tabular} & \begin{tabular}[c]{@{}c@{}}base \\ building \\ \textbf{sky} \\ square \\ \textbf{statue}\end{tabular}     && \begin{tabular}[c]{@{}c@{}}column \\ front \\ \textbf{man} \\ \textbf{sky} \\ \textbf{statue}\end{tabular} & \begin{tabular}[c]{@{}c@{}}front \\ \textbf{man} \\ \textbf{sky} \\ \textbf{statue} \\ \textbf{view}\end{tabular} \\ \bottomrule
\end{tabular}
}
\caption{Qualitative results of the baseline methods and our proposed representation on IAPR-TC12. Labels ordered according to their relevance scores.}
\label{fig:qualitative}
\end{figure*}


\subsection{Experiment 5: Scaling by Subsampling the Training Set}
\begin{figure}
\centering     
\includegraphics[width=0.49\columnwidth]{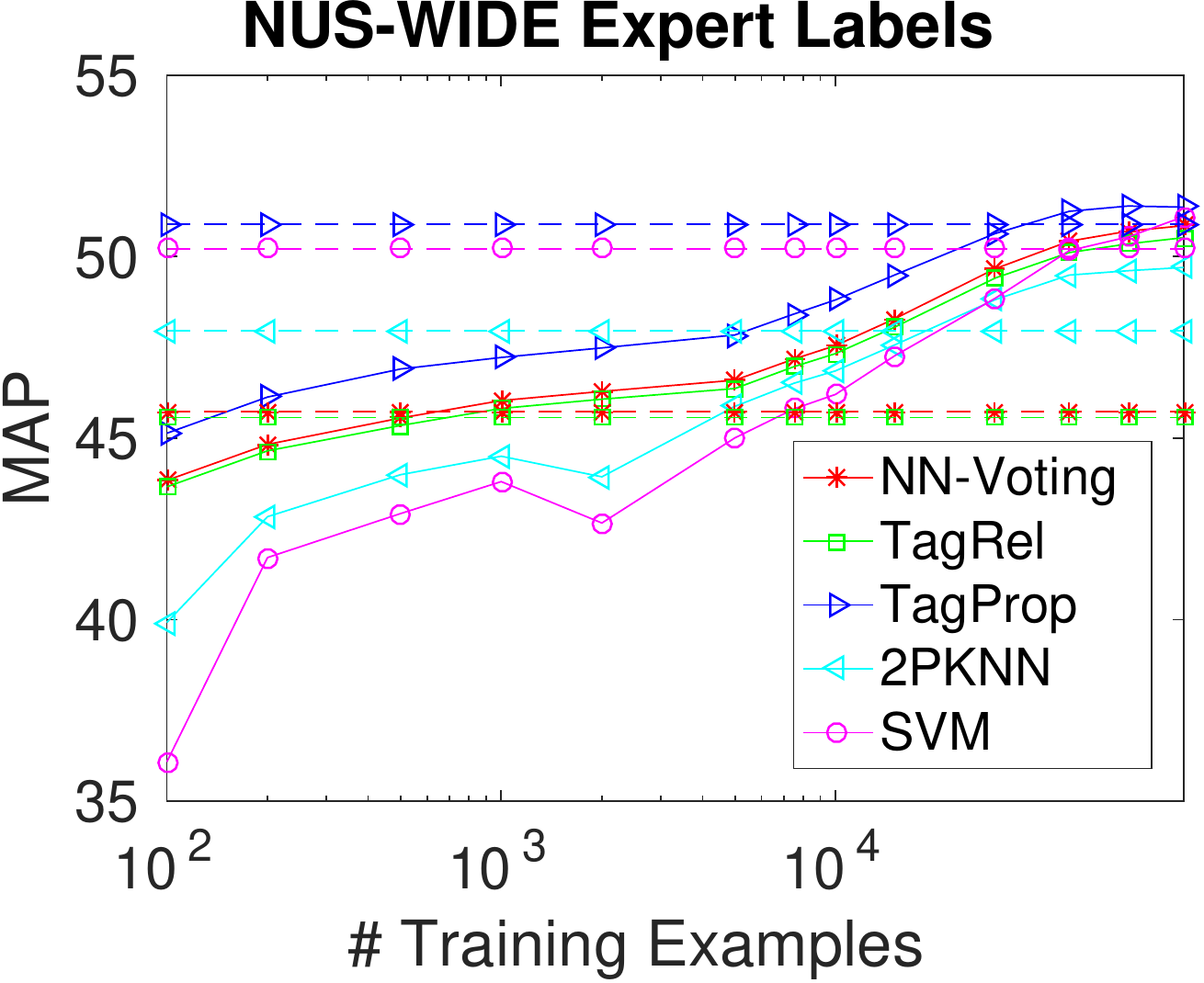}
\includegraphics[width=0.49\columnwidth]{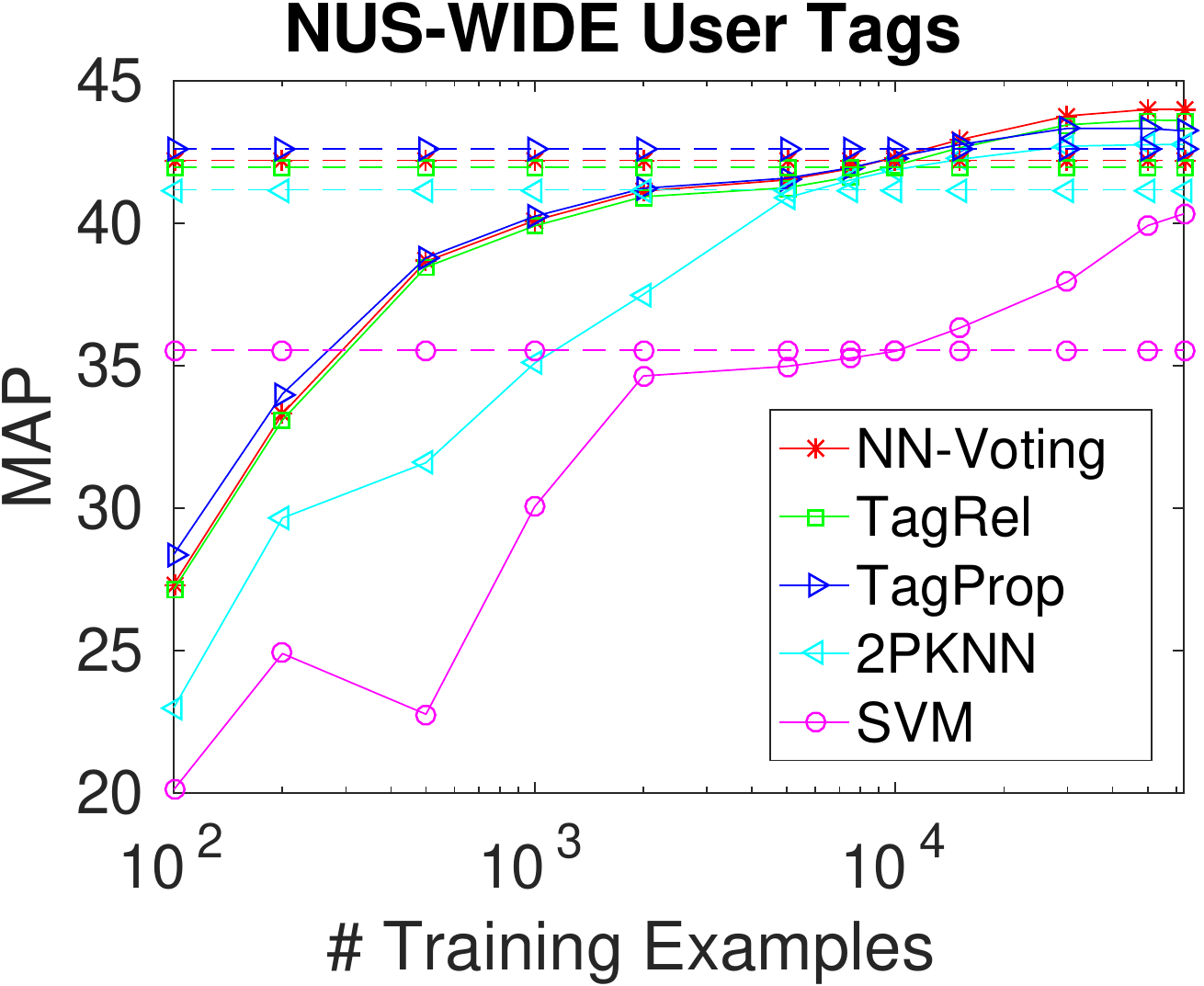}

\caption{\textbf{Training KCCA with a subset of data.} MAP of the five methods trained with KCCA on NUS-WIDE varying the number of images used for training the projections, with expert labels (on the left) and user tags (on the right). Dashed lines represent baseline methods.}
\label{fig:MAPgtsubsample}
\end{figure}

One key issue with KCCA is that it can be onerous to scale the training over millions of images.
The most expensive effort is carried out in the training phase where the projection vectors are estimated. At test time, the computational cost is negligible since it is only given by the multiplication of the features with the estimated projection vectors.

As also noted by Hardoon \etal \cite{hardoon-2004}, big training sets with large kernel matrices can lead to computational problems.
Two main issues arise: \emph{i}) high computational cost to compute the generalized eigenvalues problem, and \emph{ii}) the memory footprint of handling large kernel matrices. 

For the first issue, we compute only a reduced number of dimensions in the semantic space by using partial Gram-Schmidt orthogonalization (PGSO), i.e. we solve the generalized eigenvalues with an incomplete Cholesky factorization of the kernel matrices.
This is a reasonable approximation because the projection is built up as the span of a subset of independent projections, and it reconstructs a limited amount of energy. 

For the second issue, the memory footprint increases quadratically with the number of training images.
In this section we explore the possibility of using a subsample of the training set to manage also this problem. 
To this end, we randomly select a subset of size $M$ from the original training set used to train KCCA, and obtain the projections.
Then we use them to project the full training set and test the methods in this approximate semantic space. 
We run the experiment only on NUS-WIDE since it has the highest number of images. 
The whole experiment is repeated with five different splits in the two settings of using expert labels or user tags. Note that this setting is different from the one used in Sect.~\ref{sec:results_gt} and Sect.~\ref{sec:results_tags} for NUS-WIDE, where we used the split provided by the authors of the dataset.

Figure \ref{fig:MAPgtsubsample} shows the MAP scores obtained with a subset of the training data.
We report results by increasing $M$ from $100$ to the full training set size (with exponential steps). Using more training data, we expect the quality of the projections to be improved. 
Either with expert labels or user tags, more the training data, the better the projections obtained.
We note that a minimum quantity of data is required to obtain a performance higher than the baseline; this corresponds to the point in the figure in which the corresponding dashed and solid lines intersect each other.
The specific subset of training data depends on the method and on the quality of the annotations. 
When expert labels are available, NNvot and TagRel obtains an improvement even with a very small amount of training images. In contrast, TagProp requires more data to gain MAP because of its rank learning phase.
This means that our approach can provide some improvements even when very few labeled images are available, but more data may be needed with advanced nearest neighbors schemes.
Considering the scenario of user tags, the three methods show similar performance with similar numbers of training images.
This suggests that differently from expert labels, the noise in user tags is responsible for the hampered performance and more data is needed to reliably estimate good projections.   
\begin{figure}
\centering
\includegraphics[width=.85\columnwidth]{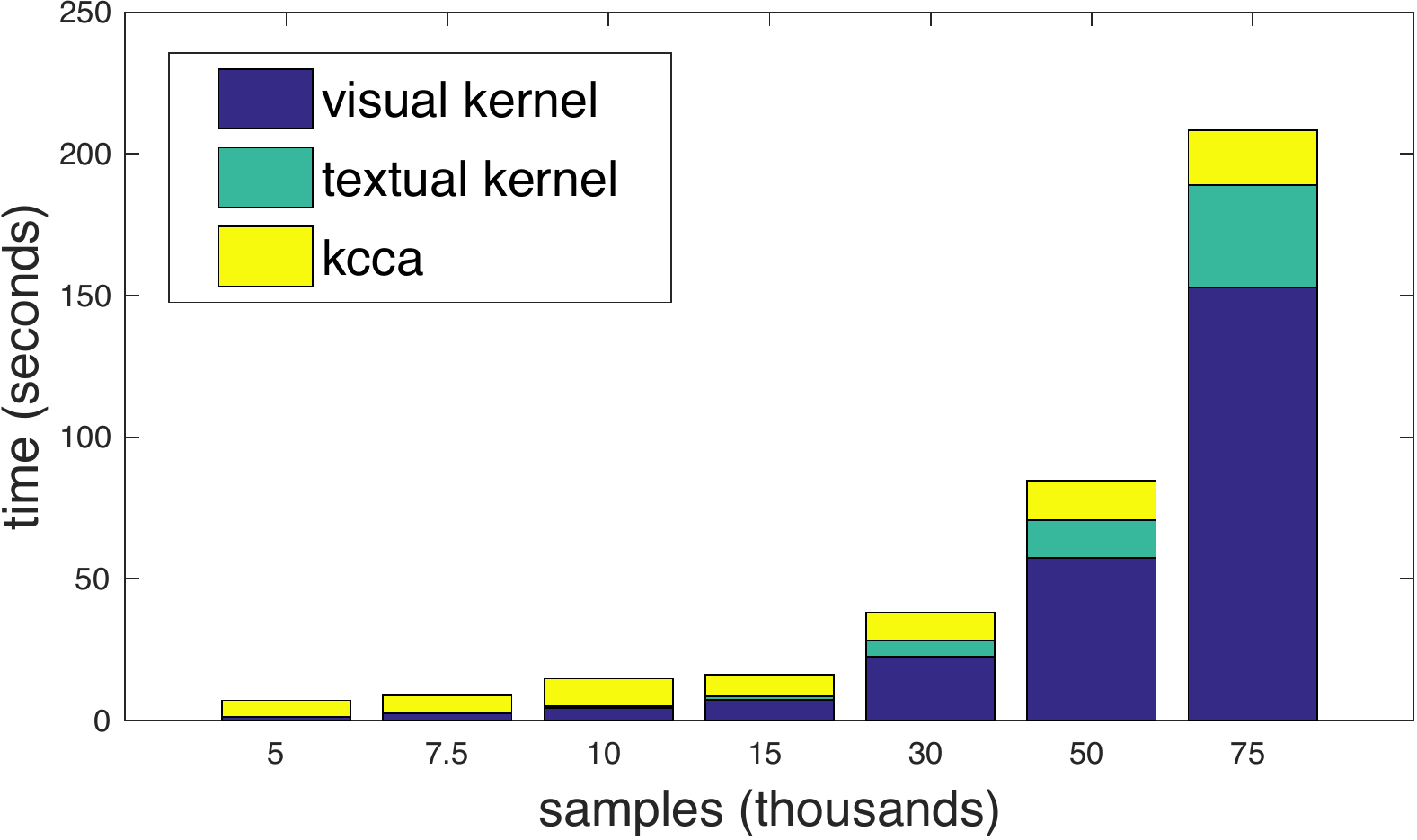}
\caption{Timing of our approach varying the number of samples employed for learning KCCA. We report separately the time for visual kernel, textual kernel and KCCA computation. The time is dominated by the visual kernel computation.\label{fig:timing}}
\end{figure}

We evaluate the additional computational cost of our approach, by timing the run of KCCA on NUS-WIDE on our sub-sampling experiment. It can be noted from Fig.~\ref{fig:timing} that the overall computation is dominated by the visual kernel computation. 
Since we approximated the kernel matrices with GSD to a fixed rank value, the running time required to compute the KCCA projections can only increase up to a fixed maximum value, independently from the number of samples.

\subsection{Qualitative Analysis}

Figure~\ref{fig:qualitative} shows four examples of annotations produced by our method on the IAPR-TC12 dataset.
It can be seen that TagProp and TagRel perform better for both baseline representation and the proposed semantic space.
Thanks to the integration of labels into the semantic space, our technique allows nearest neighbor methods to distinguish between visually similar but semantically different images.
Look for instance at the first example: a salt desert. Baseline approaches wrongly predict that this might be a \dquote{beach} image, since the salt visually resembles sand. Differently, our semantic space dismisses beach images and allows NN methods to find samples with \dquote{desert} and \dquote{salt}, thus obtaining a correct image labeling.

Moreover, our method can also deal with information that was missing in the visual space. A good example is given by the second picture shown in Figure~\ref{fig:qualitative}.
This image depicts two people and an \dquote{hammock}. Since the label \dquote{hammock} is not in the $1$K concepts used to train the VGG16 network, similar hammock images are difficult to be retrieved for the baseline methods.
In contrast, our method has integrated this missing information into the semantic space, allowing TagRel and TagProp to find semantically similar images and predict the presence of the hammock correctly.

The third and fourth images demonstrate that our technique is able to bring closer images with fine-grained labels.
For instance, the third image is a close-up of a person wearing several well visible clothing.
Baseline methods correctly found easy concepts like \dquote{man}, \dquote{cap} or \dquote{hair}, while label transfer methods operating in the semantic space can also predict more specific labels such as \dquote{shirt}, \dquote{polo} and \dquote{portrait}. 
Finally, the fourth image depicts a statue portrayed from below, in contrast with the blue sky. This image is correctly annotated with the difficult labels \dquote{man} and \dquote{view} only by TagProp when trained on the semantic space.

\section{Conclusion}\label{sec:conclusions}
This paper presents a novel automatic image annotation framework based on KCCA. Our work shows that it is indeed useful to integrate textual and visual information into a semantic space that is able to preserve correlation with the respective original features.
Our method does not require the textual information at test time, and it is therefore suitable for label prediction on unlabeled images.
We additionally propose a label denoising algorithm that allows to exploit user tags in place of expert labels. This scenario is of extreme interest given the abundance of images with user tags that can be extracted from social media.
Finally, we show that semantic projections can be learned also with a subset of the training set, making it possible to obtain some benefits even on large-scale datasets.

We report extensive experimental results on all the classic automatic image annotation datasets, as well as more recent datasets collected from Flickr. Our experiments show that label transfer in the semantic space allows consistent improvement over standard schemes that rely only on visual features. All the best performing image annotation methods have shown to be able to exploit the proposed embedding.
We believe that our framework will provide a strong baseline to compare and better understand future automatic image annotation algorithms.

\section*{Acknowledgments}
This work was supported by the MIUR project No. CTN01\_ 00034\_23154\_SMST.
L. Ballan was supported by the EU's FP7 under the grant agreement No.~623930 (Marie Curie IOF).


\section*{References}





\bibliographystyle{model1-num-names}
\bibliography{imgannot}

\end{document}